\RequirePackage{fix-cm}
\documentclass[smallextended]{svjour3}       

\smartqed  
\usepackage{graphicx}
\usepackage{amssymb}
\usepackage{geometry}
\usepackage{subfig}
\usepackage{caption}
\usepackage{algorithm}
\usepackage{algpseudocode}
\usepackage{amsmath}
\usepackage{natbib}
\usepackage{multirow}
\usepackage{makecell}
\usepackage[misc]{ifsym} 
\begin{document}

\title{Boosting-GNN: Boosting Algorithm for Graph Networks on Imbalanced Node Classification
}

\titlerunning{Boosting-GNN}        

\author{{Shuhao Shi}    \textsuperscript{1}   \and
        {Kai Qiao}   \textsuperscript{1}  \and
        {Shuai Yang}   \textsuperscript{1}  \and
        {Linyuan Wang}   \textsuperscript{1}  \and
        {Jian Chen }   \textsuperscript{1}  \and
        {Bin Yan}   \textsuperscript{1,~\Letter}
}
\institute{%
    \begin{itemize}
      \item[] {Shuhao Shi} \\
            \email{ssh$\_$smile@163.com}
      \\
           \item[]{Kai Qiao} \\
           \email{qiaokai1992@gmail.com}
      \\
           \item[]{Shuai Yang} \\
           \email{ysstation@foxmail.com}
      \\
           \item[]{Linyuan Wang} \\
           \email{wanglinyuanwly@163.com}
      \\
           \item[]{Jian Chen} \\
           \email{kronhugo@163.com}
      \\
           \item[\textsuperscript{\Letter}]{Bin Yan} \\
           \email{ybspace@hotmail.com}
      \at
      \item[\textsuperscript{1}] Henan Key Laboratory of Imaging and Intelligence Processing, PLA strategy support force information engineering university, Zhengzhou, China
    \end{itemize}
}


\date{Received: date / Accepted: date}

\maketitle

\begin{abstract}
The Graph Neural Network (GNN) has been widely used for graph data representation. However, the existing researches only consider the ideal balanced dataset, and the imbalanced dataset is rarely considered. Traditional methods such as resampling, reweighting, and synthetic samples that deal with imbalanced datasets are no longer applicable in GNN. This paper proposes an ensemble model called Boosting-GNN, which uses GNNs as the base classifiers during boosting. In Boosting-GNN, higher weights are set for the training samples that are not correctly classified by the previous classifier, thus achieving higher classification accuracy and better reliability. Besides, transfer learning is used to reduce computational cost and increase fitting ability. Experimental results indicate that the proposed Boosting-GNN model achieves better performance than GCN, GraphSAGE, GAT, SGC, N-GCN, and most advanced reweighting and resampling methods on synthetic imbalanced datasets, with an average performance improvement of 4.5\%.

\end{abstract}

\section{Introduction}
\label{intro}
Convolutional Neural Networks (CNNs) have been widely used in image recognition \citep{article29,article49}, object detection \citep{article21}, speech recognition \citep{article48}, visual coding and decoding \citep{article44,article45}. However, traditional CNNs can only handle data in Euclidean space. It cannot effectively address the heterogeneous graphs that are prevalent in real life. Graph Neural Networks (GNNs) can effectively construct deep learning models on graphs. In addition to homogeneous graphs, Heterogeneous Graph Neural Networks \citep{article51,article46,article47} can effectively handle more comprehensive information and semantically richer heterogeneous graphs.
\par The Graph Convolutional Network (GCN) \citep{article9} has achieved remarkable success in multiple graph data-related tasks, including recommendation system \citep{article23,article26}, molecular recognition \citep{article36}, traffic forecast \citep{article22}, and point cloud segmentation \citep{article27}. GCN is based on the neighborhood aggregation scheme, which generates node embedding by combining information from neighborhoods. GCN achieves superior performance in solving node classification problems compared with conventional methods, but it is adversely affected by datasets imbalance. However, existing studies on GCNs all aim at balanced datasets, and the problem of imbalanced datasets have not been considered.
\par In the field of machine learning, the processing of imbalanced data sets is always challenging \citep{article1,article13}. The data distribution of imbalanced dataset makes the model's fitting ability insufficient because it is difficult for the model to learn useful information from unevenly distributed datasets \citep{article16}. A balanced dataset consists of almost the same number of training samples in each class. In reality, it is impractical to obtain the same number of training samples for different classes because the data in different classes is generally not uniformly distributed \citep{article16,article32}. The imbalance of training dataset are caused by many possible factors, such as deviation sampling and measurement errors. Samples may be collected from narrow geographical areas in a specific time period and in different areas at different times, exhibiting a completely different sample distribution. The datasets widely used in deep learning research, \itshape{e.g.}\upshape, ImageNet ILSVRC 2012 \citep{article29}, MS COCO \citep{article21}, and Places Database \citep{article30}, \textit{etc.}\upshape, are balanced datasets, where the amount of data in different classes is basically the same. Recently, more and more imbalanced datasets reflecting real-world data characteristics have been built and released, \textit{e.g.}\upshape, iNaturalist \citep{article39}, LVIS \citep{article40}, and RPC \citep{article41}. It is difficult for traditional pattern recognition methods to achieve excellent results on imbalanced datasets, so methods that can deal with imbalanced datasets efficiently are urgently needed.
\par For imbalanced datasets, additional processing is needed to reduce the adverse effects \citep{article16}. The existing machine learning methods mainly rely on resampling, data synthesis, and reweighting. 1) Resampling samples the original data by undersampling and oversampling. Undersampling removes part of data in the majority class, so that the majority class can match with the minority class in terms of the amount of data. Oversampling copies the data in minority class. 2) Data synthesis, \textit{i.e.}\upshape, SMOTE \citep{article4} and its improved version \citep{article18,article19,article32} as well as other synthesis methods \citep{article31}, synthesize the new sample artificially by analyzing the samples in minority class. 3) Reweighting assigns different weights to different samples in the loss function to improve the model's performance on imbalanced datasets.
\par In the GNN, the existing processing methods for imbalanced datasets in machine learning is not applicable. 1) The data distribution problem of imbalanced datasets cannot be overcome by resampling. The use of oversampling may introduce many repeated samples to the model, which reduces the training speed and leads to overfitting easily. In the case of undersampling, valuable samples that are important to feature learning may be discarded, making it difficult for the model to learn the actual data distribution. 2) The use of data synthesis method or oversampling method loses the relationship between the newly generated samples and the original samples in the GNN, which affects the aggregation process of nodes. 3) Reweighting, e.g., Focal Loss \citep{article28}, CB Focal Loss \citep{article37} and \textit{etc}\upshape, can solve the problem of imbalanced dataset in GCN to some extent, but it do not consider the relationship between training samples, and fails to achieve a satisfactory performance in dealing with imbalanced datasets.
\par Ensemble learning methods are more effective in improving the classification performance of imbalanced data than data sampling techniques \citep{article43}. It is challenging for a single model to classify rare and few samples on an imbalanced dataset accurately, thus the overall performance is limited. Ensemble learning is a process of aggregating multiple base classifiers to improve the generalization ability of classifiers. Briefly, ensemble learning uses multiple weak classifiers to make classification on the dataset. In traditional machine learning, ensemble learning is used to improve the classification accuracy of multi-class imbalanced data \citep{article34,article24,article12,article11,article10,article8}. In CNNs, some models adopt ensemble learning to deal with imbalanced datasets. ERFS-based CNN \citep{article52} adaptively resamples the training set in iterations to get multiple classifiers and forms a cascade ensemble model. AdaBoost-CNN \citep{article1} integrates AdaBoost with a CNN to improve accuracy on imbalanced data.
\par Inspired by ensemble learning, an ensemble GNN classifier that can deal with the imbalanced dataset is proposed in this paper. The AdaBoost algorithm is combined with GNN to train the GNN classifier by serialization, and the samples are reweighted according to the calculation results. Based on this, the proposed classifier improving the classification performance on the imbalanced dataset.
The main contributions of this paper are as follows:
\par • This paper is the first to study the imbalanced dataset problem in GNN. An Boosting-GNN model is proposed to deal with imbalanced datasets in semi-supervised nodes classification. A transfer learning strategy is also applied to speed up training of the Boosting-GNN model.
\par • Four imbalanced datasets are constructed to evaluate performance of the Boosting-GNN. Boosting-GNN uses GCN, GAT and GraphSAGE as base classifiers, improving the classification accuracy on imbalanced datasets.
\par • The robustness of Boosting-GNN under feature noise perturbations is discussed, and it is discovered that ensemble learning can significantly improve the robustness of GNNs.
\par The rest of this paper is organized as follows. Sect. \ref{sec:1} introduces the related work of dealing with imbalanced data sets and the application of ensemble learning in deep learning. In Sect. \ref{sec:5}, the principle of the proposed Boosting-GNN is discussed. Then, four datasets and a proposed method for performance evaluation are described, and the experimental results are discussed in Sect. \ref{sec:8}. Finally, Sect. \ref{sec:17} concludes the paper.

\section{Related works}
\label{sec:1}
Due to the prevalence of imbalanced data in practical applications, the problem of imbalanced data sets has attracted more and more attention. Recent researches are mainly conducted in the following four directions:
\subsection{Resampling}
\label{sec:2}
Resampling can be specifically divided into two types: 1) Oversampling by copying data in minority classes \citep{article3,article5}. After oversampling, some samples are repeated in the dataset, leading to a less robust model and worse generalization performance on imbalanced data. 2) Undersampling by selecting data in the majority class \citep{article3,article5}. Undersampling may cause information loss in major class. The model only learns a part of the overall pattern, leading to underfitting \citep{article6}. $K$-means and stratified random sampling to undersample (KSS) \citep{article42} performs undersampling after $K$-means clustering for majority classes, and achieves good results.
\subsection{Synthetic samples}
\label{sec:3}
The data synthesis methods generate samples similar to a few samples in the original set. The representative method is SMOTE \citep{article4}, and the operations of this method is as follows. For each sample in a small sample set, an arbitrary sample is selected from its $K$-nearest neighbors. Then, a random point on the line between the sample and the selected sample is taken as a new sample. However, the overlapping degree will be increased by synthesizing the same number of new samples for each minority class. The Borderline-SMOTE \citep{article32} synthesizes new samples similar to the samples on the classification boundary. SMOTE-RSB* \citep{article18} exploits the synthetic minority oversampling technique and the editing technique based on the rough set theory. G-SMOTE \citep{article19} generates a synthesized sample for each of the selected instances in a geometric region of the input space. ADASYN \citep{article31} algorithm synthesizes different number of new samples for different minority samples.
\subsection{Reweighting}
\label{sec:4}
Reweighting typically assigns different weights to different samples in the loss function. In general, reweighting assigns large weights to training samples in minority classes \citep{article38}. Besides, finer control of loss can be achieved at the sample level. For example, Focal Loss \citep{article28} designed a weight adjustment scheme to improve the classification performance of imbalanced dataset. CB Focal Loss \citep{article37} introduced a weight factor inversely proportional to the number of effective samples to rebalance the loss, reaching the most advanced level in the imbalanced dataset.

\subsection{Ensemble classifiers}
Ensemble classifiers are more effective than sampling methods to deal with the imbalance problem \citep{article43}. In GNN models, AdaGCN \citep{article2} integrating Adaboost and GCN layers to get deeper network models. Different from AdaGCN, Boosting-GNN uses GNN as a sub-classifier of Boosting algorithm to improve the performance on imbalanced datasets. To our knowledge, we are the first to use ensemble learning to solve the classification on graph imbalanced datasets.
\par In addition, there are transfer learning, domain adaptation, and other methods to deal with imbalance problem. The method based on transfer learning solves the problem by transferring the characteristics learned from majority classes to minority classes \citep{article17}. Domain adaptive method processes different types of data and learns how to reweight adaptively \citep{article35}. These methods are beyond the scope of this article.

\section{The Proposed Method}
\label{sec:5}
\subsection{GCN model}
\label{sec:6}
Given an input undirected graph $\mathcal{G}=\left\{\mathcal{V},\mathcal{E} \right\}$, where $\mathcal{V}$ and $\mathcal{E}$ respectively denote the set of $N$ nodes and the set of $e$ edges. The corresponding adjacency matrix $A\in {{\mathbb{R}}^{N\times N}}$ is a $N\times N$ sparse matrix. The entry $(i,j)$ in the adjacency matrix is equal to 1 if there is an edge between $i$ and $j$, and 0 otherwise. The degree matrix $D$ is a diagonal matrix where each entry on the diagonal indicates the degree of vertex, which can be computed as ${{d}}_{i}=\sum\nolimits_{j}{{{a}_{ij}}}$. Each node is associated with an $F$-dimensional feature vector, and $X\in {{\mathbb{R}}^{N\times F}}$ denotes the feature matrix for all nodes.
GCN model of semi-supervised classification has two layers \citep{article9}, and every layer computes the transformation:

\begin{equation}
{{H}^{(l+1)}}=\sigma ({{Z}^{(l+1)}}),{Z}^{(l+1)}=\tilde{A}{{H}^{(l)}}{W}^{(l)}
\end{equation}

\noindent where $\tilde{A}$ is normalized adjacency obtained by $\tilde{A}={{D}^{-\frac{1}{2}}}A{{D}^{-\frac{1}{2}}}$. ${{W}^{(l)}}$ is the trainable weights of the layer. $\sigma (\cdot )$ denotes an activation function (usually ReLU), and ${{H}^{(l)}}\in {{\mathbb{R}}^{N\times {{d}_{l}}}}$ is the input activation matrix of the $\l$th hidden layer, where each row is a ${{d}_{l}}$-dimensional vector for node representation. The initial node representations are the original input features:

\begin{equation}
{{H}^{(0)}}=X
\end{equation}

A two-layer GCN model can be defined in terms of vertex features $X$ and $\hat{A}$ as:

\begin{equation}
{\rm GCN_{2-layer}}(\hat{A},X;\theta )=softmax(\hat{A}\cdot \sigma (\hat{A}X{{W}^{(0)}}){{W}^{(1)}})
\end{equation}

The GCN is trained by the back propagation learning algorithm. The last layer uses the \textit{softmax}\upshape{} function for classification, the cross-entropy loss over all labeled examples are evaluated:

\begin{equation}
\mathcal{L}=-\sum\limits_{\left| {{\mathcal{Y}}_{L}} \right|}{\sum\limits_{i\in{{\mathcal{Y}}_{L}}}loss({{y}_{i},z_{i}^{L}})}
\end{equation}

Formally, given a dataset with $n$ entities $\left( X,Y \right)=\left\{ \left( {{x}_{i}},{{y}_{i}} \right) \right\}_{i=1}^{N}$, where ${{x}_{i}}$ represents the word embedding for entity $i$, and ${{y}_{i}}\in \left\{ 1,\cdot \cdot \cdot \cdot \cdot \cdot ,C \right\}$ represents the label for ${{x}_{i}}$. Multiple weak classifiers are combined with AdaBoost algorithm to make a single strong classifier.

\subsection{Proposed Algorithm}
\label{sec:7}
Since ensemble learning is an effective method to deal with imbalanced datasets, Boosting-GNN adopts the Adaboost algorithm proposed by \citep{article7} to design an ensemble strategy for GCNs, which can train the GCNs sequentially. In Boosting-GNN, the weight of each training sample is assigned according to the degree to which the sample was not correctly trained in the previous classifier.
\paragraph{Aggregation}~{}
\newline
Boosting-GNN aggregates GNN through Adaboost algorithm to improve the performance on imbalanced datasets. First, the overall formula of Boosting-GNN can be expressed as:

\begin{equation}
\label{equ:Aggregation1}
{{F}_{M}}(x)=\sum\limits_{m=1}^{M}{{{\alpha }_{m}}}*{{G}_{m}}(x;{{\theta }_{m}})
\end{equation}

\noindent where ${{F}_{M}}(x)$ is the ensemble classifier obtained after $M$ rounds of training, and $x$ denotes samples. A new GNN classifier ${{G}_{m}}(x;{{\theta }_{m}})$ is trained in each round, and ${{\theta }_{m}}$ is the optimal parameter learned by the base classifier. The weight of the classifier ${{\alpha }_{m}}$ denotes the importance of classifier and it could be obtained according to the error of the classifier. Based on (\ref{equ:Aggregation1}), Formula (\ref{equ:Aggregation2}) can be obtained:

\begin{equation}
\label{equ:Aggregation2}
{{F}_{m}}(x)={{F}_{m-1}}(x)+{{\alpha }_{m}}*{{G}_{m}}(x;{{\theta }_{m}})
\end{equation}

${{F}_{m-1}}(x)$ is the weighted aggregation of the previously trained base classifier. In each iteration, a new base classifier ${{G}_{m}}(x;{{\theta }_{m}})$ and its weights ${{\alpha }_{m}}$ are solved. Boosting-GNN uses an exponential loss function:

\begin{equation}
L(y,F(x))={{e}^{-y*F(x)}}
\end{equation}

According to the meaning of the loss function, if the classification is correct, the exponent part is a negative number; otherwise, it is a positive number. As for training the base classifier, the training dataset is $T=\left\{ ({{x}_{i}},{{y}_{i}})_{i=1}^{N} \right\}$; ${{x}_{i}}$ is the feature vector of the $i$th node; ${{y}_{i}}$ is the category label of the $i$th node, and ${{y}_{i}}\in \left\{ 1,...,C \right\}$, where $C$ is the total number of classes.

\paragraph{Reweight Samples}~{}
\newline
Assume that during the first training, the samples are evenly distributed and all weights are the same. The data weights are initialized by ${{D}_{1}}=\left\{ w_{1}^{1},w_{2}^{1},...,w_{N}^{1} \right\}$, where $w_{i}^{1}=1/N, i=1,...,N$, and $N$ is the number of samples. Training $M$ networks in sequence on the training set, the expected loss ${{\varepsilon }_{m}}$ at the $m$th iteration is:

\begin{equation}
\label{equ:reweight1}
{{\varepsilon }_{m}}=\sum\limits_{{{y}_{i}}\ne {{G}_{m}}({{x}_{i}};{{\theta }_{m}})}{w_{i}^{m}=}\sum\limits_{i=1}^{N}{w_{i}^{m}}\mathbb I ({{y}_{i}}\ne {{G}_{m}}({{x}_{i}};{{\theta }_{m}}))
\end{equation}

\noindent where $\mathbb I$ is the indicator function. When the input is true, the function value is 1; otherwise, the function value is 0. ${{\varepsilon }_{m}}$ is the sum of the weights of all misclassified samples. ${{\alpha }_{m}}$ can be treated as a hyper-parameter to be tuned manually, or as a model parameter to be optimized automatically. In our model, to keep it simply, ${{\alpha }_{m}}$ is assigned according to ${{\varepsilon }_{m}}$.

\begin{equation}
\label{equ:reweight2}
{{\alpha }_{m}}=\frac{1}{2}\ln \frac{1-{{\varepsilon }_{m}}}{{{\varepsilon }_{m}}}
\end{equation}

${{\alpha }_{m}}$ decreases as ${{\varepsilon }_{m}}$ increases. The first GNN is trained on all the training samples with the same weight of $1/N$, indicating the same importance for all samples. After the $M$ estimators are trained, the output of GNN can be obtained, which is a $C$-dimensional vector. The vector contains the predicted values of $C$ classes, which indicate the confidence of belonging to the corresponding class. For the $m$th GNN input sample ${{x}_{i}}$, the output vector is ${{p}^{m}}({{x}_{i}})$. $p_{k}^{m}({{x}_{i}})$ is the $k$th element of ${{p}^{m}}({{x}_{i}})$, where $k=1,2,\cdot \cdot \cdot ,C$.

\begin{equation}
\label{equ:reweight3}
w_{i}^{m+1}=w_{i}^{m}{{e}^{\left( -a\frac{C-1}{C}{{y}_{i}}\log \left( {{p}^{m}}({{x}_{i}}) \right) \right)}}
\end{equation}

$w_{i}^{m}$ is the weight of the $i$th training sample of the $m$th GNN. ${{y}_{i}}$ is the one-hot label vector encoded according to the $i$th training sample. Formula (\ref{equ:reweight3}) is obtained based on Adaboost's Samme.r algorithm \citep{article7}, which is used to update the weight of sample. If the output vector of the misclassified sample is not related to the output label, a large value is obtained for the exponential term, and the misclassified sample will be assigned a larger weight in the next GNN classifier. Similarly, a correctly classified sample will be assigned a smaller weight in the next GNN classifier. In summary, the weight vector $D$ is updated so that the weight of the correctly classified samples is reduce and the weight of the misclassified samples is increased.
\par After the weights of all training samples for the current GNN are updated, they are normalized by the sum of weights of all samples. When the classifier ${{F}_{m}}(x)$ is trained, the weight distribution of the training dataset is updated for the next iteration. When the subsequent GNN-based classifier is trained, the GNN training does not start from a random initial condition. Instead, the parameters learned from the previous GNN are transferred to the $(m+1)$th GNN, so GNN is fine-tuned based on the previous GNN parameters. The use of transfer learning can reduce the number of training epochs and make the model fit faster.

Moreover, due to the change of weight, the subsequent GNN focuses on untrained samples. The subsequent GNN performs training from scratch on a small number of training samples, which easily causes overfit. For a large number of training samples, the expected label output ${{p}^{m}}({{x}_{i}})$ by the GNN after training has a strong correlation with the real label ${{y}_{i}}$. For the subsequent GNN classifier, the trained samples have a smaller weight than the sample without previous GNN training.

\paragraph{Testing with Boosting-GNN}~{}
\newline
After training the $M$ base classifiers, Equation (\ref{equ:test1}) can be used to predict the category of the input sample. The outputs of $M$ base classifiers are summed. In the summed probability vector, the category with the highest confidence is regarded as the predicted category.

\begin{equation}
\label{equ:test1}
Q(x)=\underset{k}{argmax}\,\sum\limits_{m=1}^{M}{h_{k}^{m}(x)}
\end{equation}

$h_{k}^{m}$ is the classification result of the $k$th sample made by the $m$th basis classifier, $k=1,2,\cdot \cdot \cdot ,C$, which can be calculated from the Equation (\ref{equ:test2}).

\begin{equation}
\label{equ:test2}
h_{k}^{m}=(C-1)\cdot \left( \log \left( p_{k}^{m}(x) \right)-\frac{1}{C}\sum\limits_{i=1}^{C}{\log \left( p_{i}^{m}(x) \right)} \right)
\end{equation}

\begin{figure*}[h]
\centering
  \includegraphics[width=0.75\textwidth]{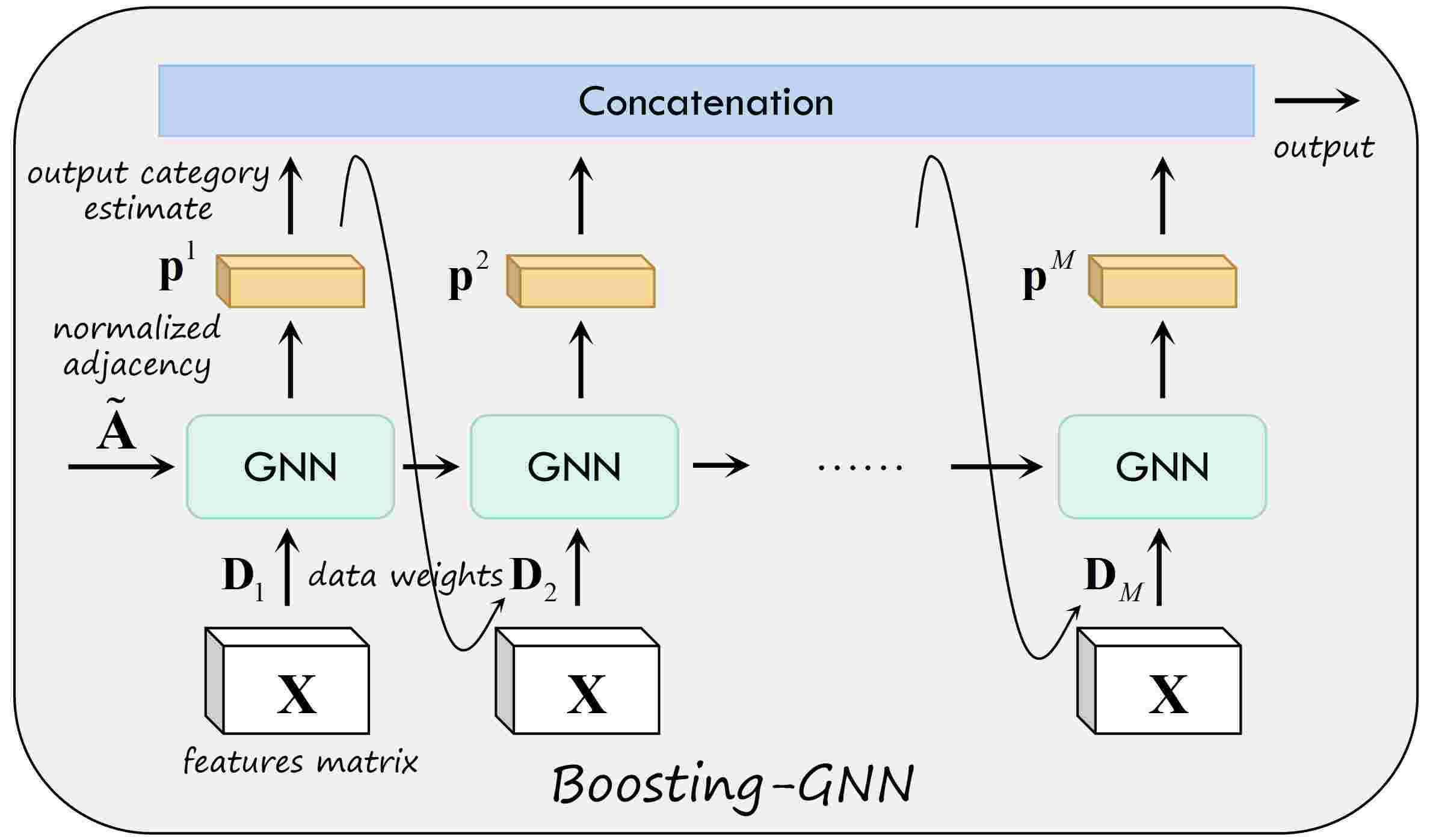}
\caption{Schematic of the proposed Boosting-GNN.}
\label{fig:1}       
\end{figure*}

Where $p_{i}^{m}(x)$ is the $k$th element of the output vector of the $m$th GCN classifier for the input $x$. Fig. \ref{fig:1} shows the schematic of the proposed Boosting-GNN. The first GNN is first trained with the initial weight ${{D}_{1}}$. Then, based on the output of the first GNN, the data weight ${{D}_{2}}$ used to update the second GNN are obtained. In addition, the parameters learned from the first GNN are transferred to the second GNN. After the $m$th base classifier is trained in order, all base classifiers are aggregated to obtain the final Boosting-GNN classifier.

The pseudo-code for an Boosting-GNN is exhibited in Algorithm \ref{alg:Framwork}. In each iteration of sequential learning, the classifiers are first trained with corresponding training data and weights. Then, according to the training results of the classifiers, the data weights are updated for the next iteration. Both operations are performed until $M$ base classifiers are trained.

\renewcommand{\algorithmicrequire}{\textbf{Input:}}  
\renewcommand{\algorithmicensure}{\textbf{Output:}} 

\begin{algorithm}[htb]
  \caption{Framework of the Boosting-GNN algorithm.}
  \label{alg:Framwork}
  \begin{algorithmic}[1]
    \Require
      Training set $T=\left\{({{x}_{1}},{{y}_{1}}),...,({{x}_{N}},{{y}_{N}}) \right\}$;
    \Ensure
      Ensemble of classifiers ${{F}_{M}}(x)$;
    \State Initialization: $w_{i}^{1}=1/N$ for all $1\le i\le N$
    \For{$m=1,2,\cdots ,N$};
    \label{code:fram:extract}
        \If {$m=1$}
            \State Train GNN classifier with weighted sample set $\left\{T,{{D}_{1}} \right\}$;
        \Else
            \State Transfer the learning parameters of the $(m-1)$th GNN to the $m$th GNN classifier;
            \State Train the $m$th GNN classifier with weighted sample set;
        \EndIf
      \State Calculate the output category estimated for the $C$ classes of the $m$th GNN classifier $p_{k}^{m}(x)$, where $k=1,2,\cdots ,C$;
      \State Calculate the training error ${{\varepsilon }_{m}}$ of the $m$th classifier according to (\ref{equ:reweight1});
      \State Assign the weight ${{\alpha }_{m}}$ to the classifier based on ${{\varepsilon }_{m}}$ using (\ref{equ:reweight2});
      \State Update the sample weight ${{D}_{m+1}}$ according to $p_{k}^{m}(x)$, and normalize the sample weight ${{D}_{m+1}}$;

    \EndFor
  \end{algorithmic}
\end{algorithm}

\section{Experiments and Analysis}
\label{sec:8}
\subsection{Experimental Settings}
\label{sec:9}
The proposed ensemble model is evaluated on three well-known citation network datasets prepared by \citep{article9}: Cora, Citeseer, and Pubmed \citep{article25}. These datasets are chosen because they are available online and are used by our baselines. In addition, experiments are also conducted on the Never Ending Language Learning (NELL) dataset \citep{article13}. As a bipartite graph dataset extracted from a knowledge graph, NELL has a larger scale than the citation datasets, and it has 210 node classes.
\paragraph{Citation networks}~{}
\newline
The nodes in the citation datasets represent articles in different fields, and the labels of nodes represent the corresponding journal where the articles published. The edges between two nodes represent the reference relationship between articles. If an edge exists between the nodes, there is a reference relationship between the articles. Each node has a one-hot vector corresponding to the keywords of the article. The task of categorization is to classify the domain of unlabeled articles based on a subset of tagged nodes and references to all articles.
\paragraph{NELL}~{}
\newline
The pre-processing schemes described in \citep{article20} are adopted in this paper. Each relationship is represented as a triplet $\left( {{e}_{1}},r,{{e}_{2}} \right)$, where ${{e}_{1}}$, $r$, and ${{e}_{2}}$ respectively represent the head entity, the relationship, and the tail entity. Each entity $E$ is regraded as a node in the graph, and each relationship $r$ consists of two nodes ${{r}_{1}}$ and ${{r}_{2}}$ in the graph.
For each $\left( {{e}_{1}},r,{{e}_{2}} \right)$, two edges $\left( {{e}_{1}},{{r}_{1}} \right)$ and $\left( {{e}_{2}},{{r}_{2}} \right)$ are added to the graph. A binary, symmetric adjacency matrix from this graph is constructed by setting entries ${{A}_{ij}}=1$, if one or more edges are present between nodes $i$ and $j$ \citep{article9}. All entity nodes are described by sparse feature vectors with dimension of 5414. Table \ref{tab:1} summarizes the statistics of these datasets.

\begin{table}[h]
\centering
\caption{Datasets used for experiments.}
\label{tab:1}       
\begin{tabular}{ccccc}
\hline\noalign{\smallskip}
Dataset & Cora & Citeseer & Pubmed & NELL  \\
\noalign{\smallskip}\hline\noalign{\smallskip}
Vertices & 2,708 & 3,327 & 19,717 &65,755 \\
Edges    & 5,429 & 4,732 & 44,338 &266,144\\
Classes  & 7     & 6     & 3      &210    \\
Features & 1,433 & 3,703 & 500    &5,414  \\
\noalign{\smallskip}\hline
\end{tabular}
\end{table}

\paragraph{Synthetic imbalanced Datasets}~{}
\newline
Different Synthetic imbalanced datasets are constructed based on the datasets mentioned above. According to the Pareto Principle that 80\% of the consequences come from 20\% of the causes, one of the classes is randomly selected as the majority category for simplicity. The remaining classes are regraded as minority classes. In \citep{article9}, 20 samples of each class were selected as the training set, and to keep the number of training samples broadly consistent, the datasets are described in the equation \ref{equ:Syndata}.

\begin{equation}
\label{equ:Syndata}
n_{i}=\left\{
 \begin{matrix}
   30 & i=c   \\
   s  & i \neq c  \\
\end{matrix} \\
 \right.
\end{equation}

${{n}_{i}}$ is the number of samples in category $i$, $c$ is the randomly selected category, $C$ is the number of classes in the dataset, and $s$ is the number of samples in the minority category. By changing $s$, the number of minority category samples is altered, thus changing the degree of imbalance in the training set. For example, in the Cora dataset, there are seven classes of samples. So, the number of samples in one class is fixed to 30, and the number of samples of the other six classes is changed. Each time the training is conducted, a certain number of samples are randomly selected to form the training set. The test set is divided following the method in \citep{article9} to evaluate the performance of different models.

\begin{figure}[htbp]
\centering
\subfloat[]{
\begin{minipage}[t]{0.3\linewidth}
\centering
\includegraphics[width=1.5in]{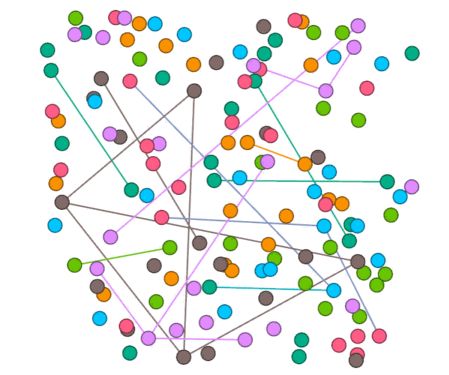}
\label{fig:syn-1}
\end{minipage}%
}
\subfloat[]{
\begin{minipage}[t]{0.3\linewidth}
\centering
\includegraphics[width=1.5in]{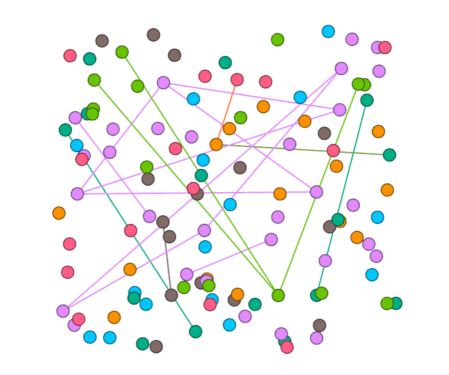}
\label{fig:syn-2}
\end{minipage}
}
\subfloat[]{
\begin{minipage}[t]{0.3\linewidth}
\centering
\includegraphics[width=1.5in]{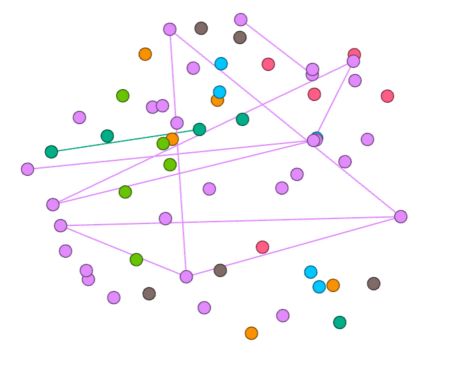}
\label{fig:syn-4}
\end{minipage}%
}
\caption{Visualisation of synthetic imbalance datasets. (a) shows the classical Cora training set. (b) shows the training set when $s$ is fetched 15. (c) is the training set when $s$ is fetched 5. The mean degrees of the nodes in (a), (b), (c) are 0.30, 0.30, 0.37 respectively.}
\label{fig:syn}
\centering
\end{figure}

Synthetic imbalanced datasets are constructed by Node Dropping. Given the graph $\mathcal{G}$, node dropping will randomly discard vertices along with their connections until the number of different classes of nodes matches the setting. In Node Dropping, each node’s dropping probability follows a uniform distribution. We visualize the synthetic datasets in Fig. \ref{fig:syn} and use different colors to represent different categories of nodes. Due to the sparsity of the adjacency matrix of the graph data set, imbalanced sampling of the graph data does not reduce the average degree of the nodes. Although disconnect parts of the graph, missing part of vertices does not affect the semantic meaning of $\mathcal{G}$.

\paragraph{Parameter Settings}~{}
\newline
In Boosting-GNN, five GNN base classifiers are used. Boosting-GNN respectively uses GCN, GraphSAGE, GAT as the base classifiers. All networks is composed of two layers, and all models are trained for a maximum of 100 epochs (training iterations) using Adam optimizer. For Cora, Citeseer and Pubmed datasets, the number of hidden units is 16, and L2 regularization is 5e-4. For NELL, the number of hidden units is 128, and L2 regularization is 1e-5.
\par The following sets of hyperparameters are used for Boosting-GNN:
For Boosting-GCN, the activation function is ReLU. The learning rates on Cora, Citeseer, Pubmed and NELL are 1e-2, 1e-2, 1e-2, 5e-3, respectively.
For Boosting-GraphSAGE the activation function is ReLU. The sampled sizes (S1 = 25, S2 = 10) is used for each layer. The learning rates on Cora, Citeseer, Pubmed and NELL are 1e-3, 1e-3, 5e-4, 1e-4, respectively.
For Boosting-GAT, the first-layer activation function is \textit{ELU}\upshape{} and the second-layer activation function is \textit{softmax}\upshape{}. The number of attention heads $K$ is 8. The learning rates on Cora, Citeseer, Pubmed and NELL are 1e-3, 1e-3, 1e-3, 5e-4, respectively.

For GCN, GraphSAGE, GAT, SGC, N-GCN and other algorithms, the models are trained for a total of 500 epochs. The highest accuracy is taken as the result of a single experiment, and the mean accuracy of ten runs with random sample split initializations is taken as the final result. A different random seed is used for every run (i.e., removing different nodes), but the ten random seeds are the same across models. All the experiments are conducted on a machine equipped with two NVIDIA Tesla V100 GPU (32 GB memory), 20-core Intel Xeon CPU (2.20 GHz), and 192 GB of RAM.

\subsection{Baseline Methods}
\label{sec:10}
The performance of the proposed method is evaluated and compared to that of three groups of methods:
\paragraph{GCN methods}~{}
\newline
In experiments, our BoostingGNN model is compared with the following representative baselines:
\par • GCN \citep{article9} produces node embedding vectors by truncating the Chebyshev polynomial to the first-order neighborhoods.
\par • GAT \citep{article14} generates node embedding vectors for each node by introducing an attention mechanism when computing node and its neighboring nodes.
\par • GraphSAGE \citep{article15} generates the embedding vector of the target vertex by learning a function that aggregates neighboring vertices. The default settings of sampled sizes (S1 = 25, S2 = 10) is used for each layer in GraphSAGE.
\par • SGC \citep{article50} reduces model complexity by eliminating the non-linearity between GCN layers, transforming a non-linear GCN into a simple linear model that is more efficient than GCNs and other GNN models for many tasks.
\par • N-GCN \citep{article33} obtains the feature representation of nodes by convolving in the neighborhood of nodes at different scales and then fusing all the convolution results. These methods can be regarded as ensemble models.

\paragraph{RS method}~{}
\newline
The KSS \citep{article42} method is used for performance comparison. KSS is a kind of $K$-means clustering method based on undersampling and achieves state-of-the-art performance on an imbalanced medical dataset.

\paragraph{RE method}~{}
\newline
Boosting-GNN is compared with GCN, GraphSAGE, and GAT. These classic models use Focal Loss \citep{article28} and CB-Focal \citep{article37}, and achieve good classification accuracy on imbalanced datasets.

\subsection{Node Classification Accuracy}
\label{sec:11}
Our method is implemented in Keras. For the other methods, the code from the Github pages introduced in the original papers is used. For synthetic imbalanced datasets, $s$ is set to 10. The classification accuracy of GCN, GraphSAGE, GAT, SGC, N-GCN and Boosting-GNN method is listed in the Table \ref{tab:2}.

\begin{table}[h]
\centering
\caption{Summary of results in terms of classification accuracy (in percentage).}
\label{tab:2}       
\begin{tabular}{ccccc}
\hline\noalign{\smallskip}
Model & Cora & Citeseer & Pubmed & NELL  \\
\noalign{\smallskip}\hline\noalign{\smallskip}
GCN         &65.6$\pm$0.8 &62.2$\pm$0.5 &71.8$\pm$0.6 &68.5$\pm$1.4\\
GraphSAGE   &66.3$\pm$0.8 &59.7$\pm$0.6 &69.7$\pm$0.6 &69.6$\pm$1.3\\
GAT         &67.4$\pm$0.7 &60.3$\pm$0.6 &66.2$\pm$0.7 &70.3$\pm$1.6\\
N-GCN       &67.3$\pm$0.6 &65.4$\pm$0.3 &72.3$\pm$0.3 &73.3$\pm$1.2\\
SGC         &69.7$\pm$0.8 &59.4$\pm$0.5 &66.9$\pm$0.5 &67.1$\pm$1.4\\
\noalign{\smallskip}\hline\noalign{\smallskip}
GCN-FL      &67.8$\pm$1.2 &65.1$\pm$0.8 &72.4$\pm$0.8 &71.2$\pm$1.2\\
GraphSAGE-FL&66.5$\pm$1.2 &59.5$\pm$0.8 &69.7$\pm$1.3 &72.1$\pm$1.1\\
GAT-FL      &67.4$\pm$1.3 &61.3$\pm$0.7 &69.2$\pm$1.2 &72.6$\pm$1.0\\
\noalign{\smallskip}\hline\noalign{\smallskip}
GCN-CB      &70.6$\pm$0.9 &65.1$\pm$0.6 &72.3$\pm$0.8 &72.9$\pm$1.4\\
GraphSAGE-CB&66.3$\pm$0.9 &59.7$\pm$0.9 &70.1$\pm$0.9 &69.8$\pm$1.4\\
GAT-CB      &67.6$\pm$1.0 &60.3$\pm$1.0 &69.3$\pm$0.9 &73.4$\pm$1.5\\
\noalign{\smallskip}\hline\noalign{\smallskip}
GCN-RS      &70.4$\pm$1.0 &61.8$\pm$1.1 &70.4$\pm$1.1 &68.9$\pm$2.1\\
\noalign{\smallskip}\hline\noalign{\smallskip}
Boosting-GCN  &73.2$\pm$0.7& \textbf{65.7$\pm$0.7} &\textbf{73.1$\pm$0.7} &74.9$\pm$1.0\\
Boosting-GraphSAGE  &72.4$\pm$1.0 &63.2$\pm$1.0 &70.4$\pm$1.1 &75.3$\pm$1.2\\
Boosting-GAT  &\textbf{73.5 $\pm$0.5} &64.3$\pm$0.8 &69.7$\pm$0.7& \textbf{75.5$\pm$1.0}\\
\noalign{\smallskip}\hline
\end{tabular}
\end{table}

Results in Table \ref{tab:2} show that Boosting-GNN outperforms the classic GNN models and state-of-the-art methods for processing imbalanced datasets. The N-GCN obtains a feature representation of the nodes by convolving around the nodes at different scales and then fusing all the convolution results, which can slightly improve the classification compared to the GCN. RS method and RE method can improve the accuracy of GNN on imbalanced datasets, but the improvement is very limited. Since RS is not suitable for graph data, RE is slightly better than RS. Boosting-GNN can significantly improve the classification accuracy of GNN, with an average increase of 6.6\%, 3.7\%, 1.8\% and 5.8\% compared with the original GNN model in Cora, Citeseer, Pubmed and NELL datasets, respectively.

Implementation details are as follows:
Following the method in \citep{article9}, 500 nodes are used as the validation set and 1000 nodes as the test set. Besides, for a fair performance comparison, the same training procedure is used for all the models.

\subsection{Effect of different levels of imbalance in the training data}
\label{sec:12}
\begin{figure}[h]
\begin{minipage}{0.24\linewidth}
\vspace{3pt}
\centerline{\includegraphics[width=\textwidth]{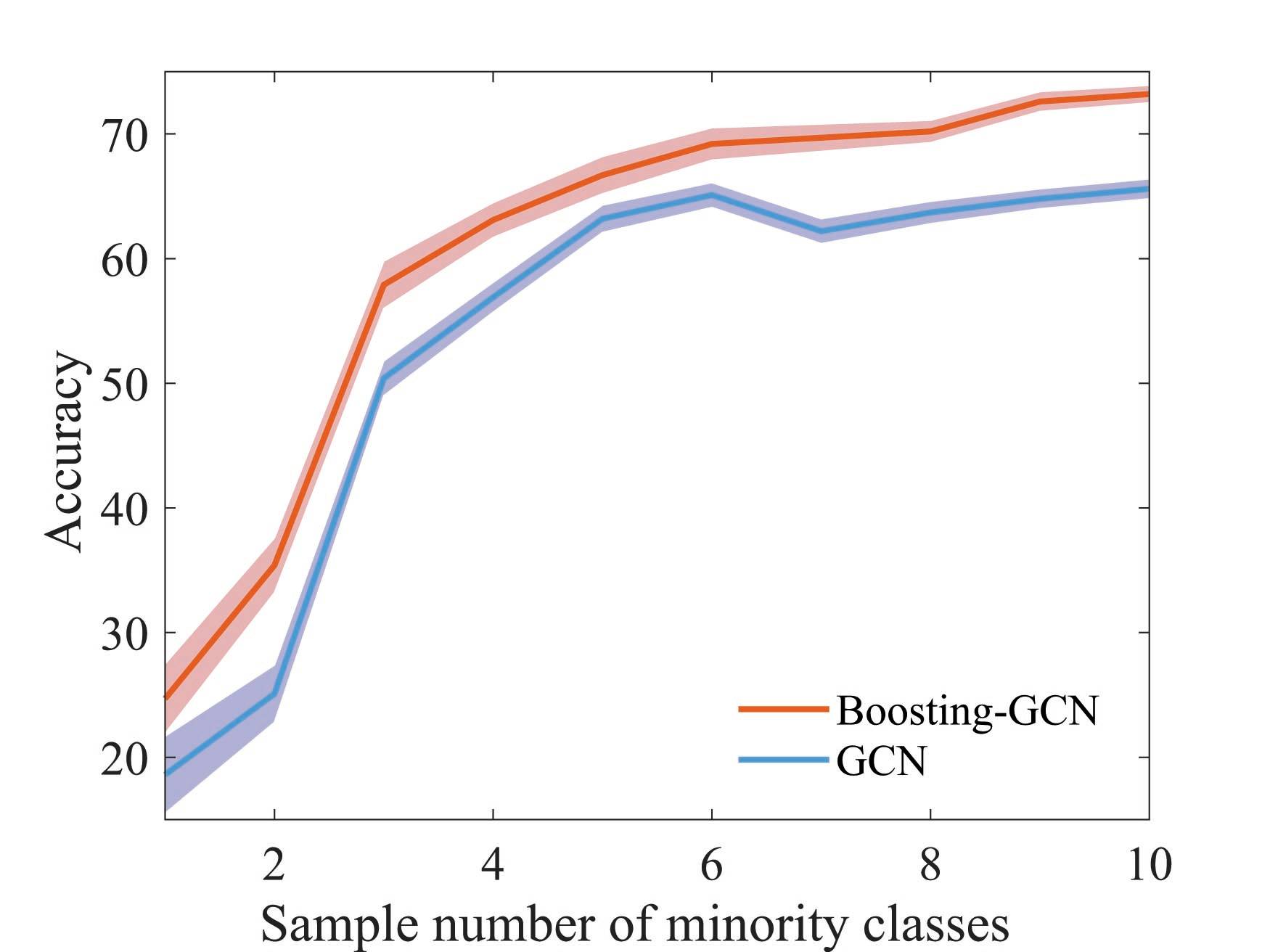}}
\vspace{3pt}
\centerline{\includegraphics[width=\textwidth]{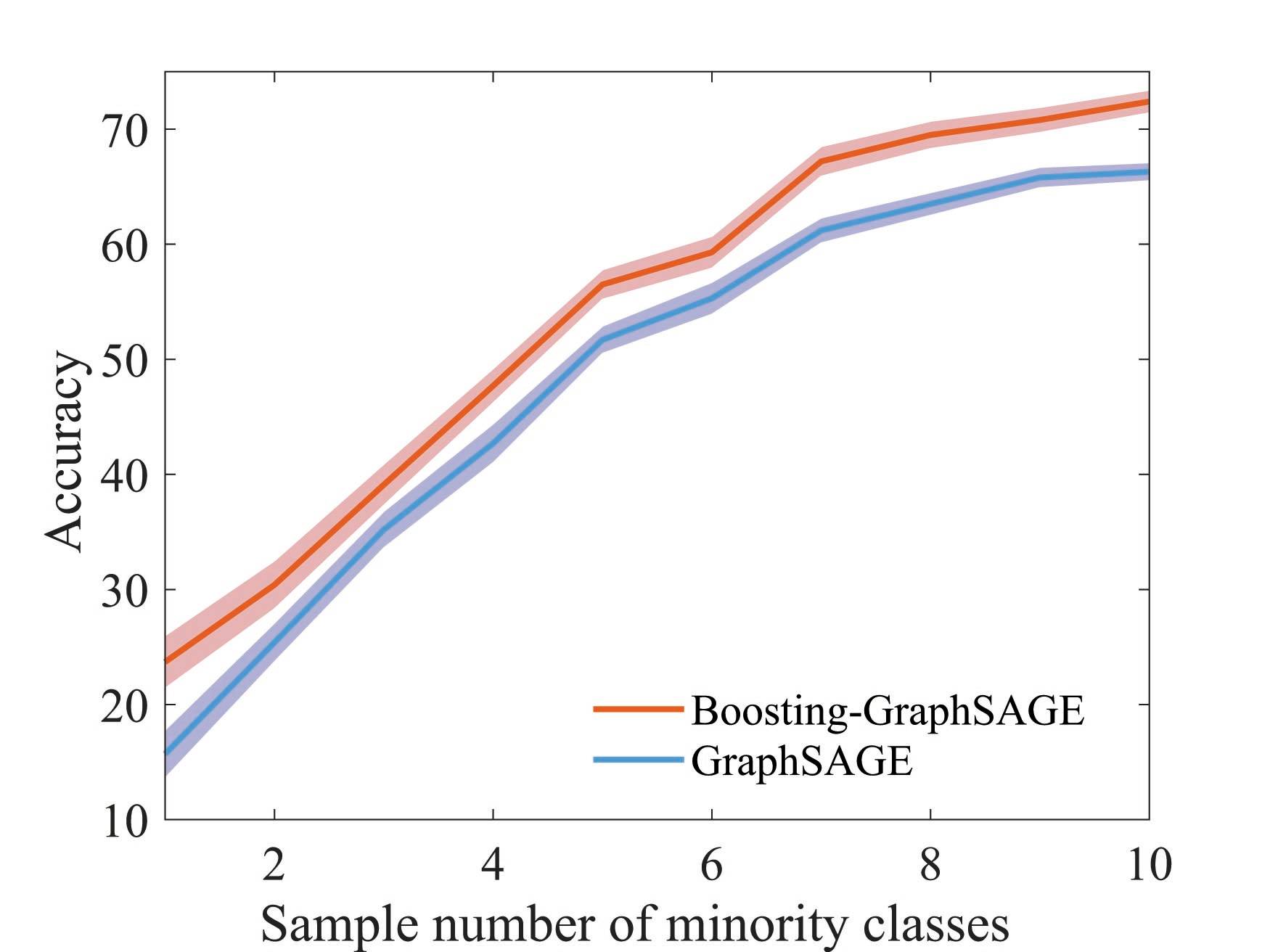}}
\vspace{3pt}
\centerline{\includegraphics[width=\textwidth]{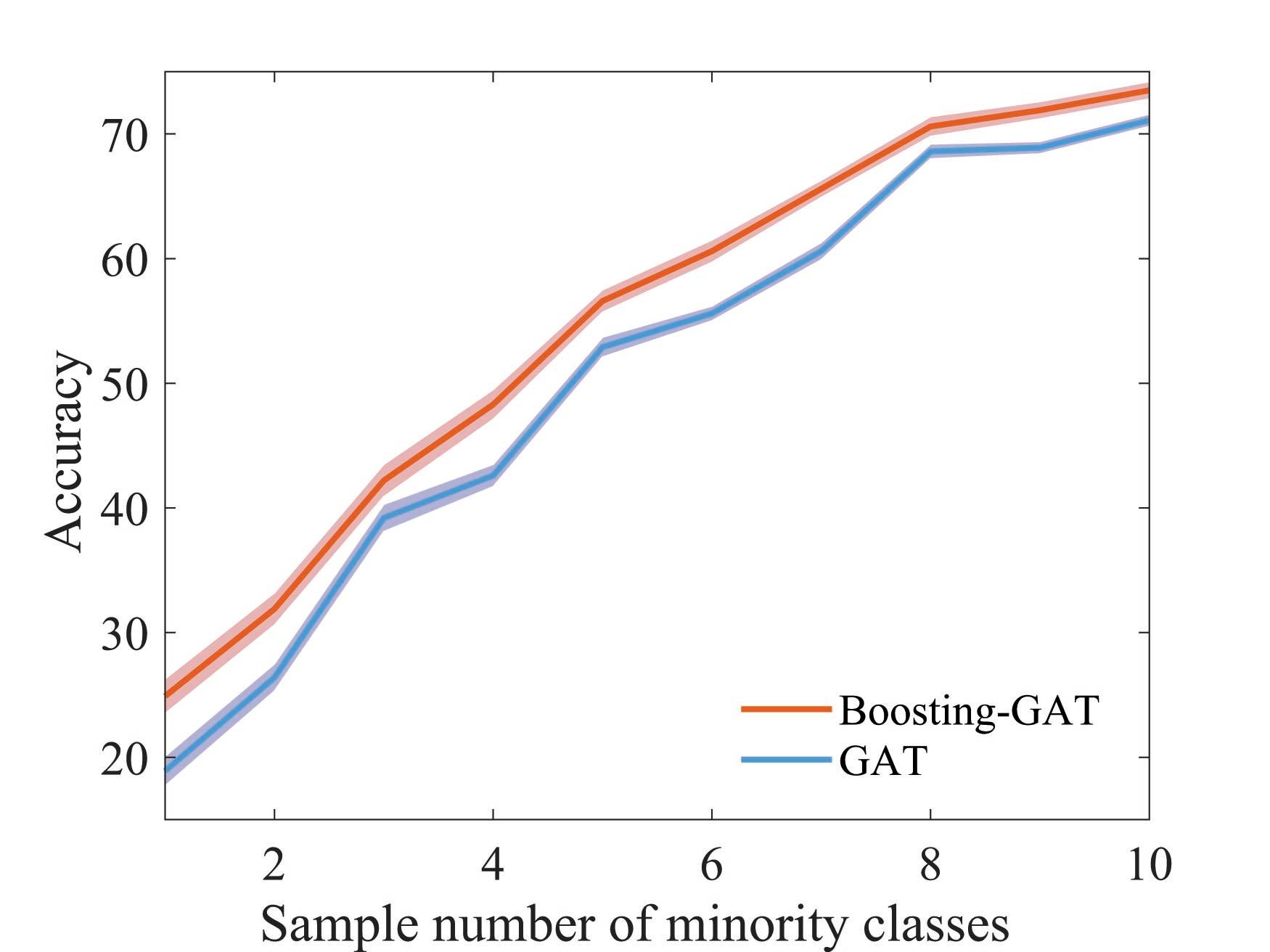}}
\vspace{3pt}
\centerline{Cora}
\end{minipage}
\begin{minipage}{0.24\linewidth}
\vspace{3pt}
\centerline{\includegraphics[width=\textwidth]{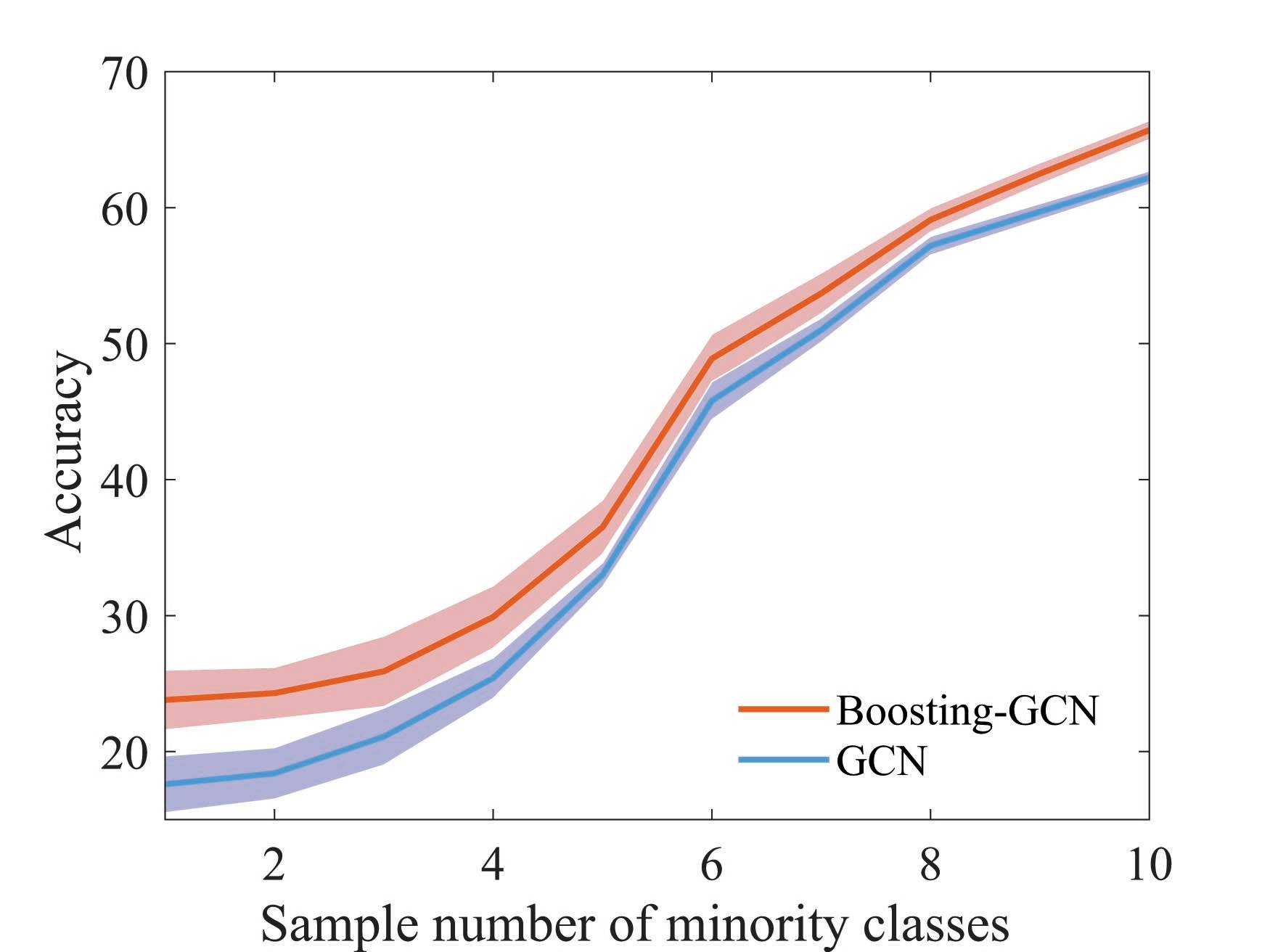}}
\vspace{3pt}
\centerline{\includegraphics[width=\textwidth]{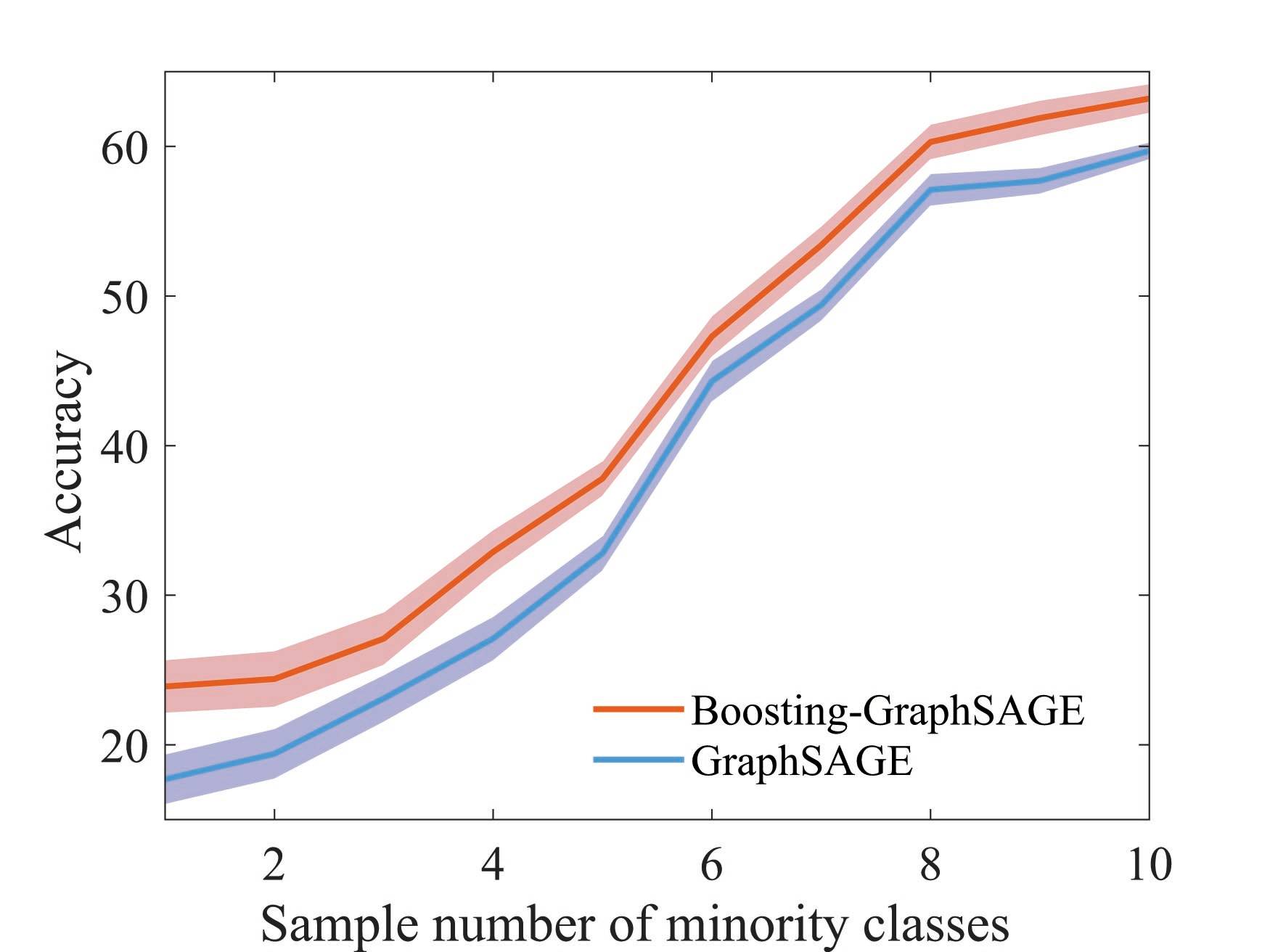}}
\vspace{3pt}
\centerline{\includegraphics[width=\textwidth]{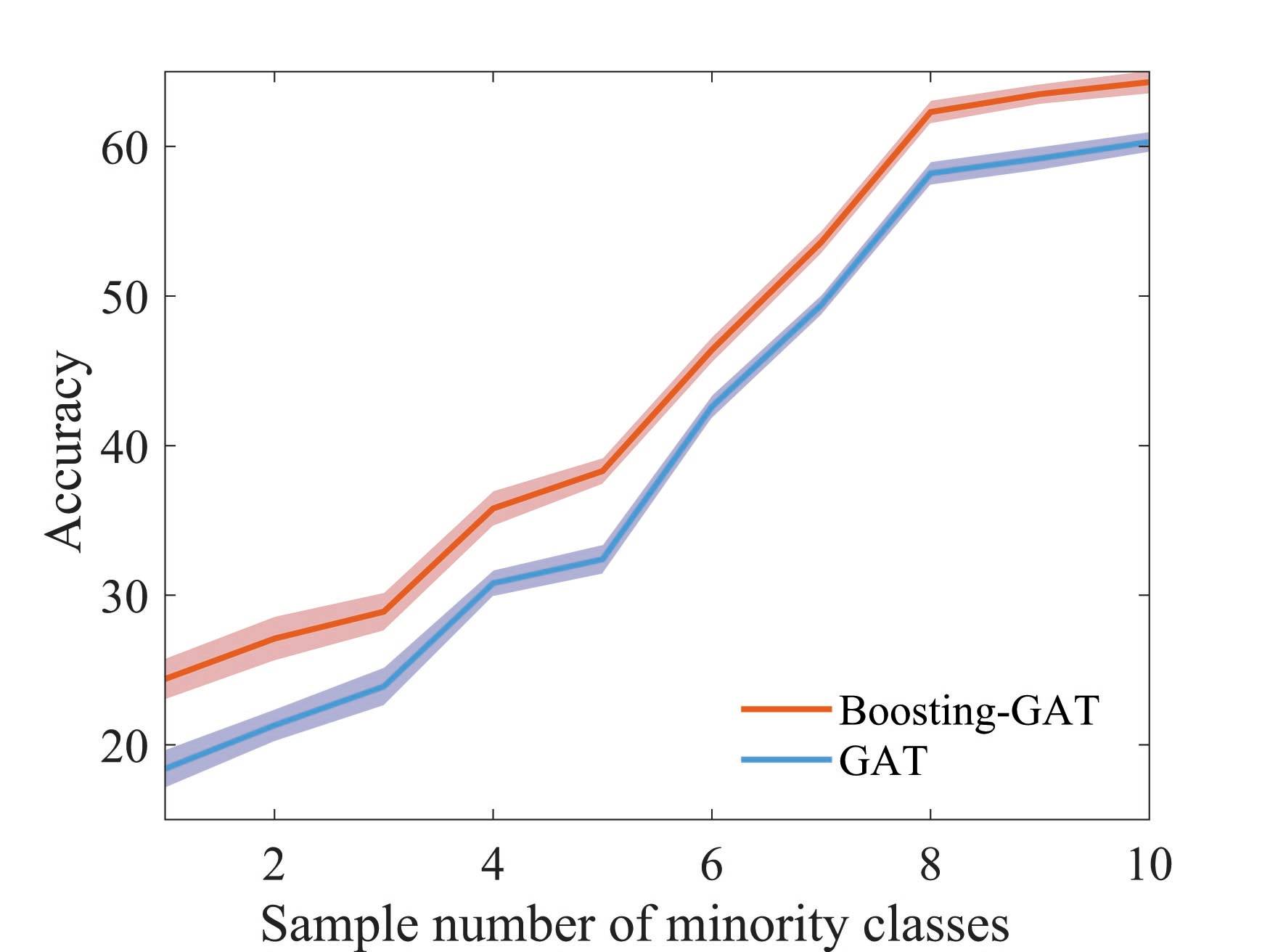}}
\vspace{3pt}
\centerline{Citeseer}
\end{minipage}
\begin{minipage}{0.24\linewidth}
\vspace{3pt}
\centerline{\includegraphics[width=\textwidth]{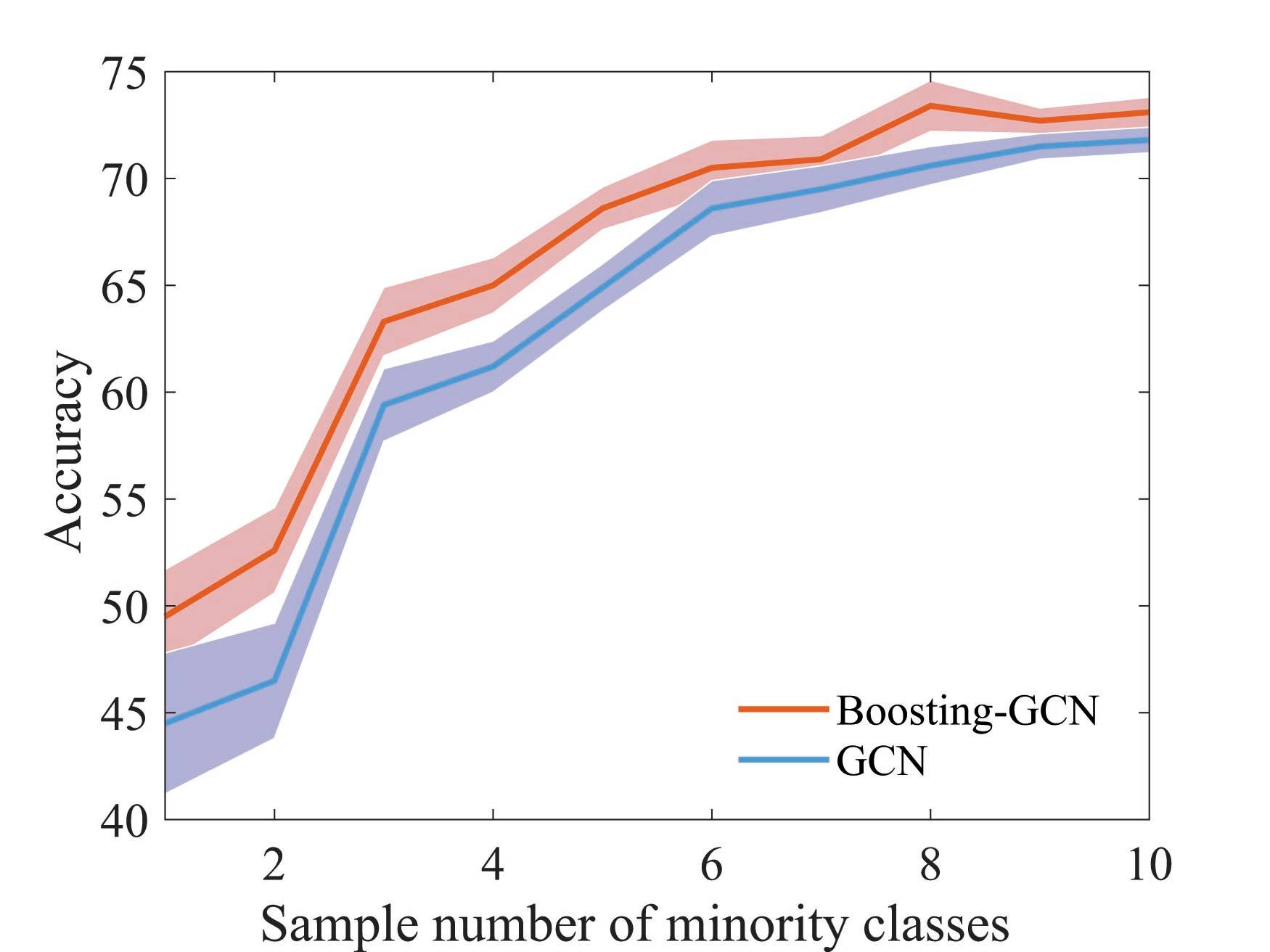}}
\vspace{3pt}
\centerline{\includegraphics[width=\textwidth]{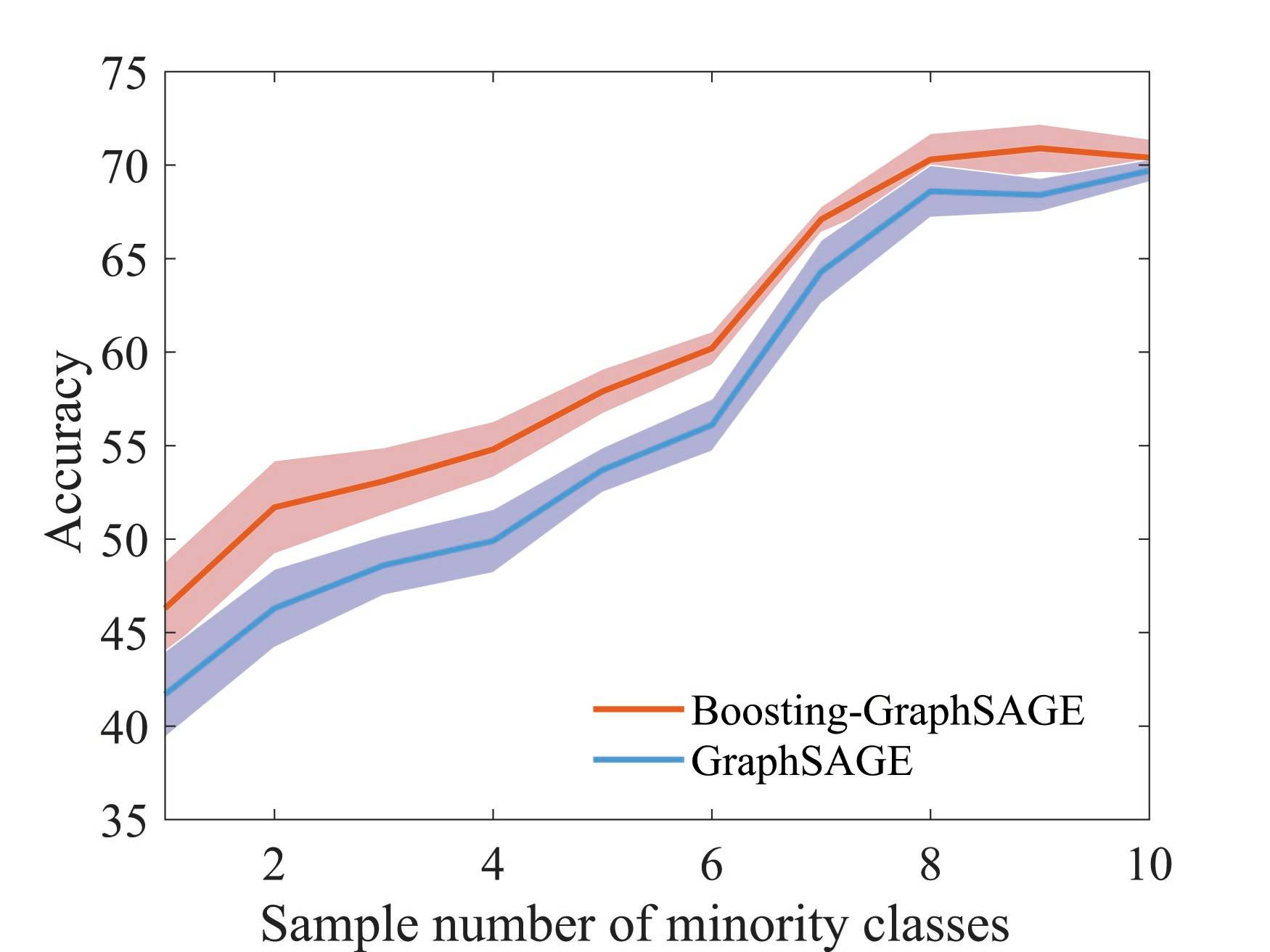}}
\vspace{3pt}
\centerline{\includegraphics[width=\textwidth]{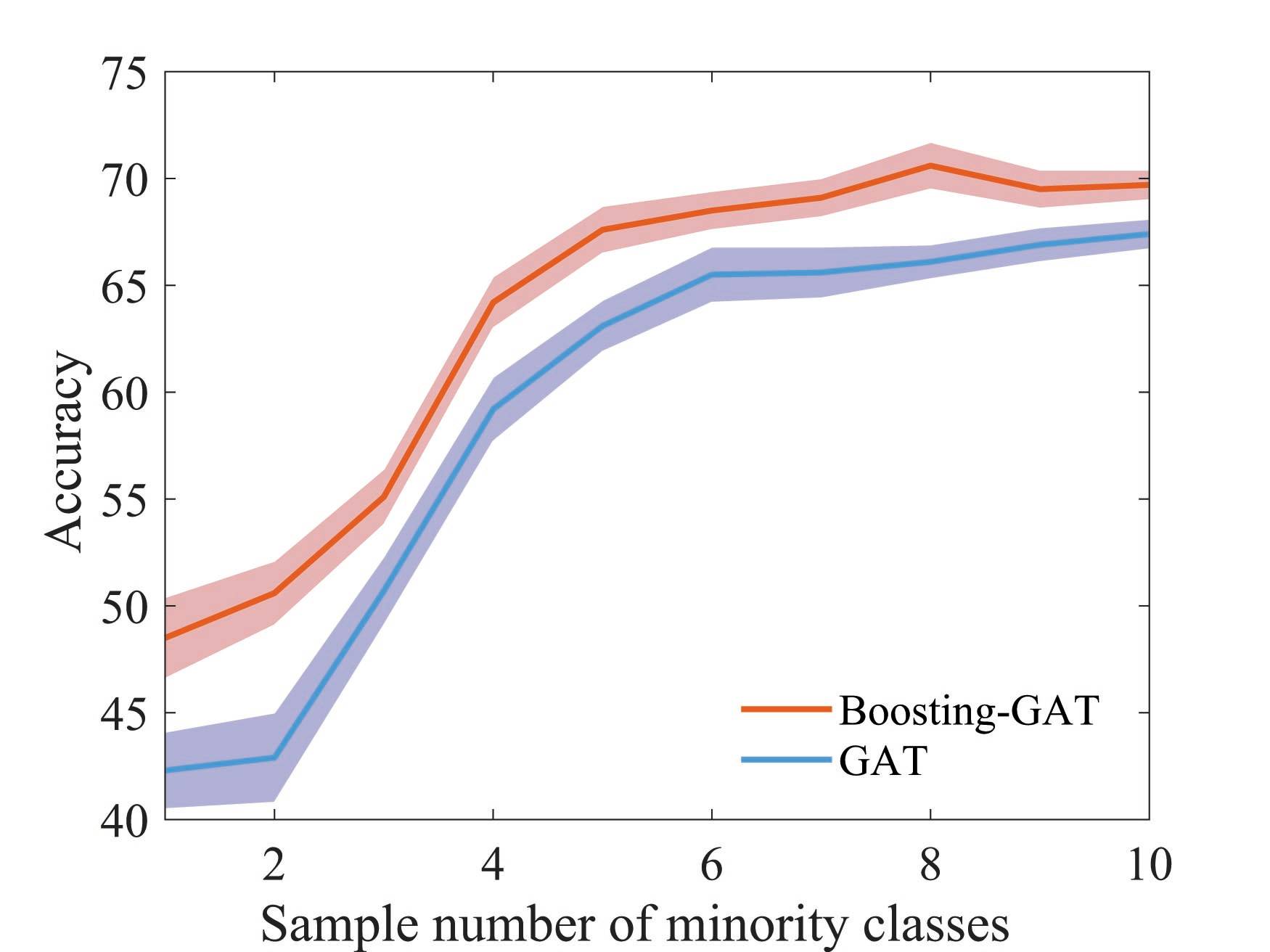}}
\vspace{3pt}
\centerline{Pubmed}
\end{minipage}
\begin{minipage}{0.24\linewidth}
\vspace{3pt}
\centerline{\includegraphics[width=\textwidth]{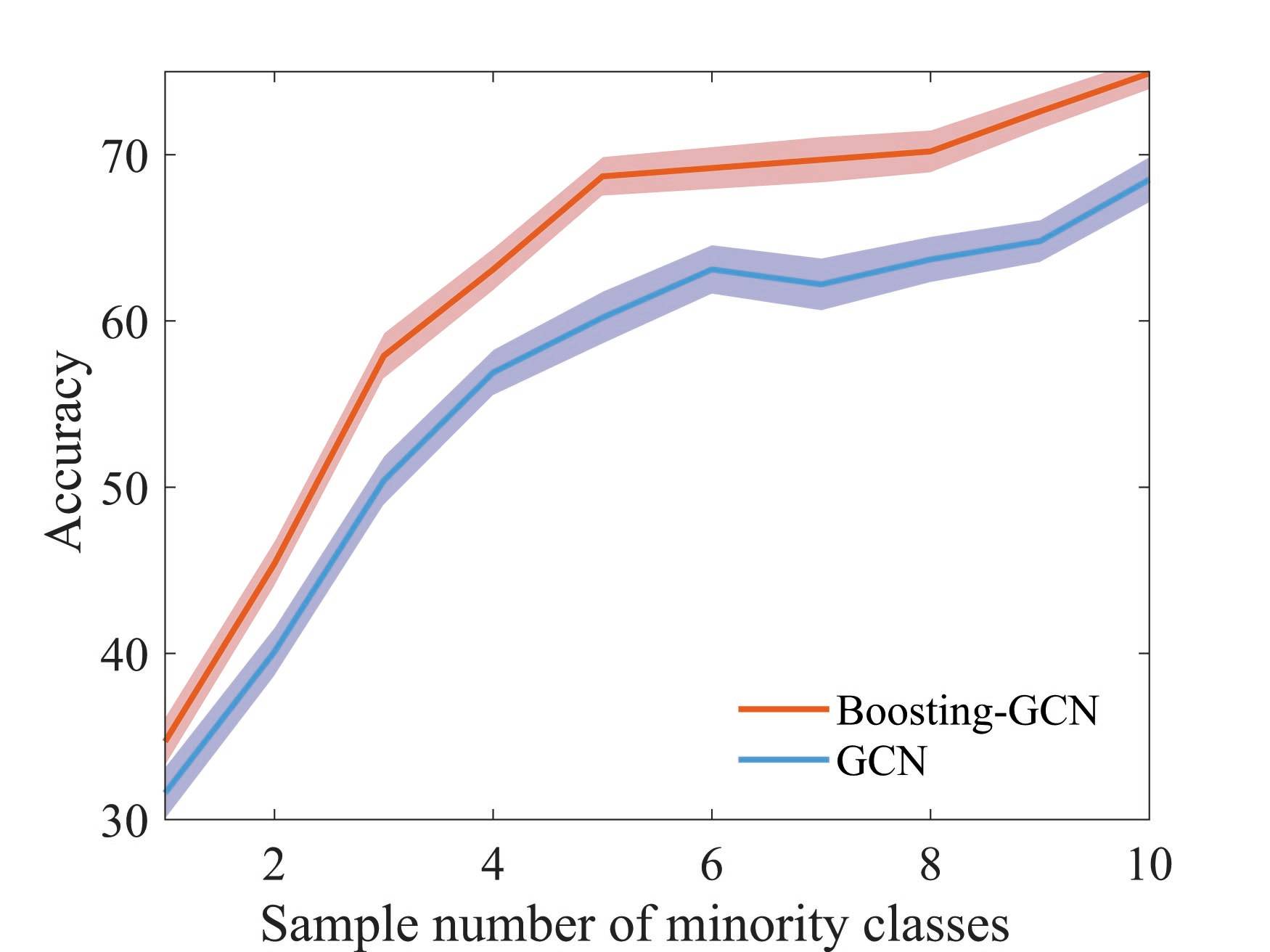}}
\vspace{3pt}
\centerline{\includegraphics[width=\textwidth]{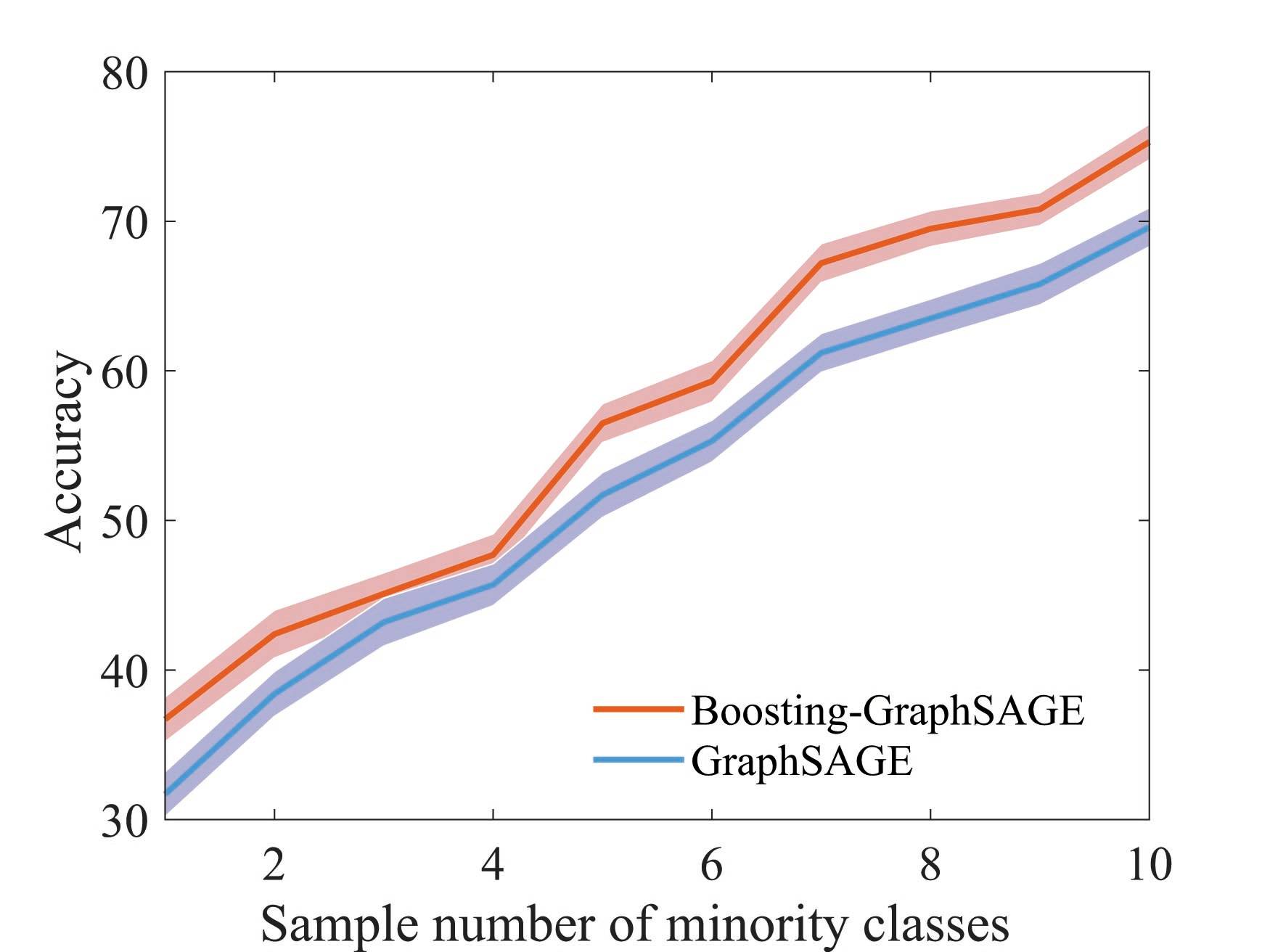}}
\vspace{3pt}
\centerline{\includegraphics[width=\textwidth]{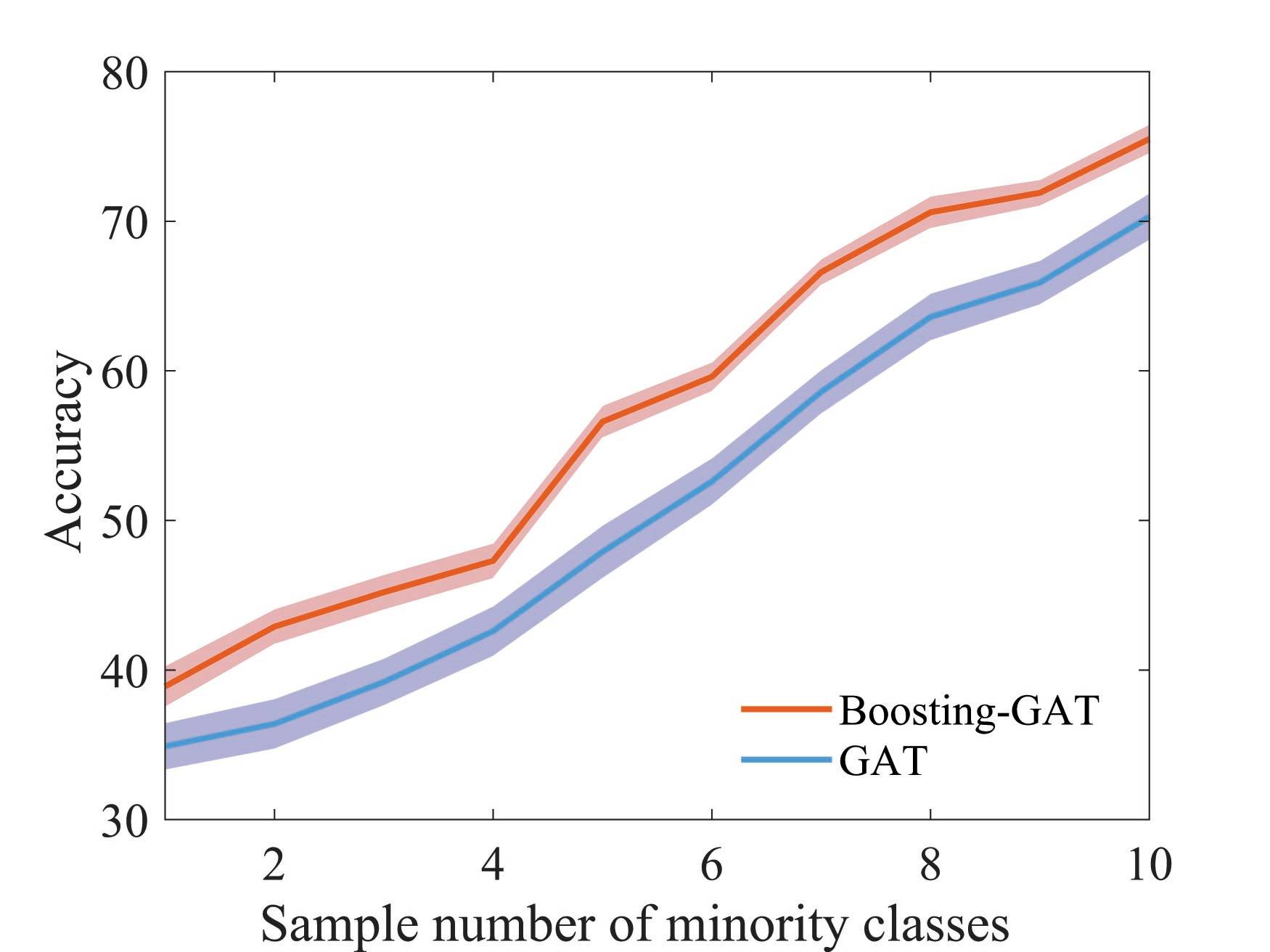}}
\vspace{3pt}
\centerline{NELL}
\end{minipage}
\caption{The classification accuracy of Boosting-GNN, GCN, GraphSAGE, and GAT on imbalanced datasets.}
\label{fig:2}
\centering
\end{figure}

The level of imbalance in the training data is changed by gradually increasing $s$ from 1 to 10. The evaluation results of Boosting-GNN, GCN, GraphSAGE, and GAT are compared, which is shown in Fig. \ref{fig:2}.

Results in Fig. \ref{fig:2} show that classification accuracy of different models varies with $s$. The shadows indicate the range of fluctuations in the experimental results. When $s$ is relatively small, the degree of imbalance in the training data is large. In this case, the classification accuracy of Boosting-GNN is higher than that of GCN, GraphSAGE, and GAT. As $s$ decreases, the performance advantage of Boosting-GNN increases gradually. Experimental results show that when the sample imbalance is large, aggregation can significantly reduce the adverse effects caused by sample imbalance and improve the classification accuracy. On Cora dataset, the accuracy of Boosting-GCN, Boosting-GraphSAGE, Boosting-GAT exceeds GCN, GraphSAGE, and GAT by 10.3\%, 8.0\%, and 6.1\% respectively at most.

\subsection{Impact of numbers of base classifiers}
\label{sec:13}
The number of base classifiers is changed to evaluate the classification accuracy on imbalanced datasets with different base classifiers. We compare the classification results of Boosting-GCN and GCN, and the experimental results is listed in Table \ref{tab:3}.
\par The experimental results show that aggregation can contribute to performance improvements. As the number of base classifiers increases, the performance improvement is more and more significant. As the number of base classifiers increases from 3 to 11, the number of base classifiers is odd. The data of Cora, Pubmed, and Citeseer are verified, and the division of train set and test set is the same as that of Sect.~\ref{sec:11}. Ten experiments are conducted, and each base classifier are trained with 100 epochs and 200 epochs. The training samples are randomly selected for each experiment.

\begin{table}[h]
\centering
\caption{Results of Boosting-GCN with varying numbers of base classifiers in terms of accuracy (in percentage).}
\label{tab:3}
\begin{tabular}{c|ccc|ccc}
\hline\noalign{\smallskip}
\multirow{2}*{\makecell{Numbers of \\base classifiers}}&
\multicolumn{3}{c|}{epoch:100}&\multicolumn{3}{c}{epoch:200} \\
   &Cora & Citeseer & Pubmed & Cora  & Citeseer & Pubmed \\ \hline
3 &\textbf{75.7$\pm$2.4}&65.5$\pm$2.5 &63.9$\pm$2.4 &75.4$\pm$2.1&65.6$\pm$1.1  &72.0$\pm$0.8\\
5 &73.2$\pm$0.7 &\textbf{65.7$\pm$0.7} &73.1$\pm$0.7 &\textbf{75.6$\pm$2.3}&\textbf{65.9$\pm$0.5}  &73.1$\pm$1.1\\
7 &73.5$\pm$1.4 &64.5$\pm$0.5 &\textbf{73.5$\pm$1.4} &74.1$\pm$2.7&64.7$\pm$0.4  &\textbf{73.5$\pm$0.8}\\
9 &72.0$\pm$0.5&63.6$\pm$0.5 &72.0$\pm$0.5 &73.9$\pm$2.0&64.2$\pm$0.3  &72.6$\pm$1.1\\
11&73.0$\pm$0.7&64.5$\pm$0.6&73.0$\pm$0.7&74.1$\pm$2.3&65.1$\pm$0.3&71.5$\pm$0.7\\
\noalign{\smallskip}\hline
\end{tabular}
\end{table}

To sum up, when the number of base classifiers is small, the classification accuracy increases with the number of base classifiers. When the number of base classifiers reaches a certain degree, the accuracy decreases due to overfitting.

\subsection{Tolerance to feature noise}
\label{sec:14}
The proposed method is tested under feature noise perturbations by removing node features randomly as \citep{article33}. This test is practical, because in the Citation networks datasets, features could be missing as article authors might forget to include relevant terms in the article abstract. By removing different features from a node, the performance of Boosting-GNN, GCN, GraghSAGE, and GAT is compared.

\begin{figure*}[h]
\centering
  \includegraphics[width=0.8\textwidth]{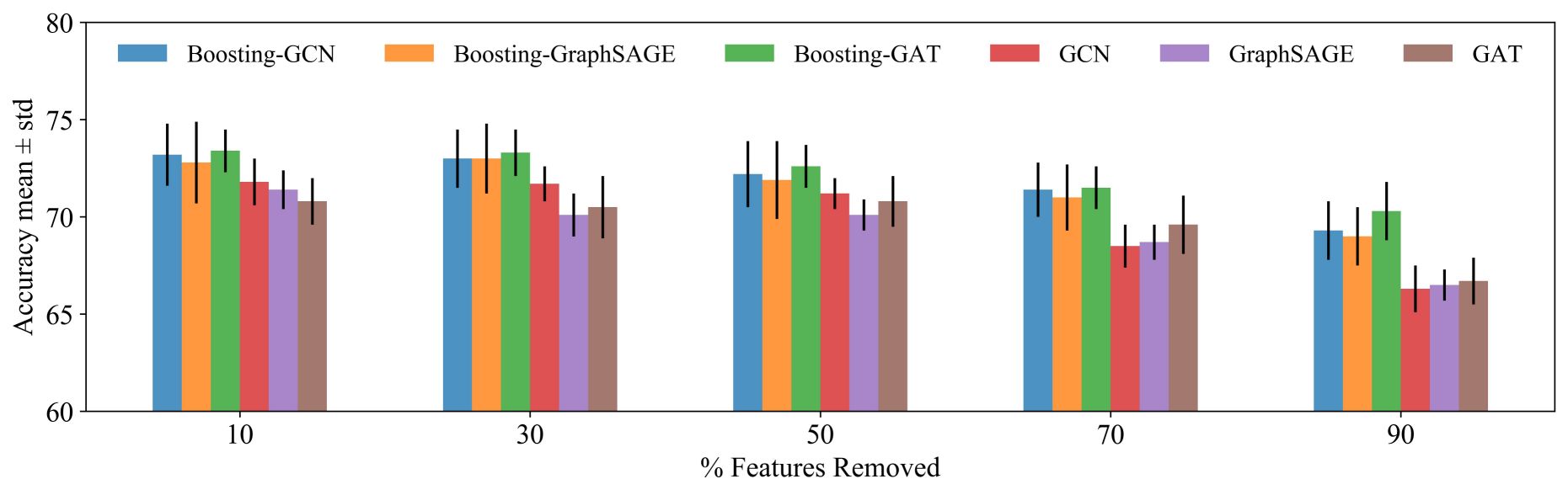}
\caption{Classification accuracy for the Cora dataset. The features are removed randomly, and the result of ten runs are averaged. A different random seed is used for every run (i.e. removing different features from each node), but the same ten random seeds are used across models.}
\label{fig:3}
\centering
\end{figure*}

Fig. \ref{fig:3} shows the performance of different methods when features are removed. As the number of removed features is increased, Boosting-GNN achieves better performance than GCN, GraghSAGE, and GAT. The greater the proportion of features removed, the greater the performance advantage of Boosting-GNN over other models. This suggests that our approach can restore the deleted features to some extent by pulling in the features directly from nearby and distant neighbors.

\subsection{Why ensemble method useful?}
\label{sec:15}
This section analyzes why the ensemble learning approach works on imbalanced datasets and the advantages of Boosting-GNN over traditional GNN.
The process of ensemble learning can be divided into two steps:
\par 1) Generating multiple base classifiers for integration. Our model could adjust the weight of samples, adopt specific strategies to reconstruct the dataset, and assign smaller weights to the determined samples and larger weights to the uncertain samples. It makes subsequent base classifiers focus more on samples that are difficult to be classified. In general, the samples of minority classes in imbalanced datasets are more likely to be misclassified. By changing the weights of these samples, subsequent base classifiers can focus more on these samples.
\par 2) Combining the results of the base classifiers. The weight of the classifier is obtained according to the error of the classifier. The base classifier with high classification accuracy has greater weight and a greater influence on the final combined classifier. In contrast, the base classifier with low classification accuracy has less weight and impact on the final combined classifier.

\par
\begin{figure}[htbp]
\centering
\subfloat[Cora]{
\begin{minipage}[t]{0.3\linewidth}
\centering
\includegraphics[width=1.5in]{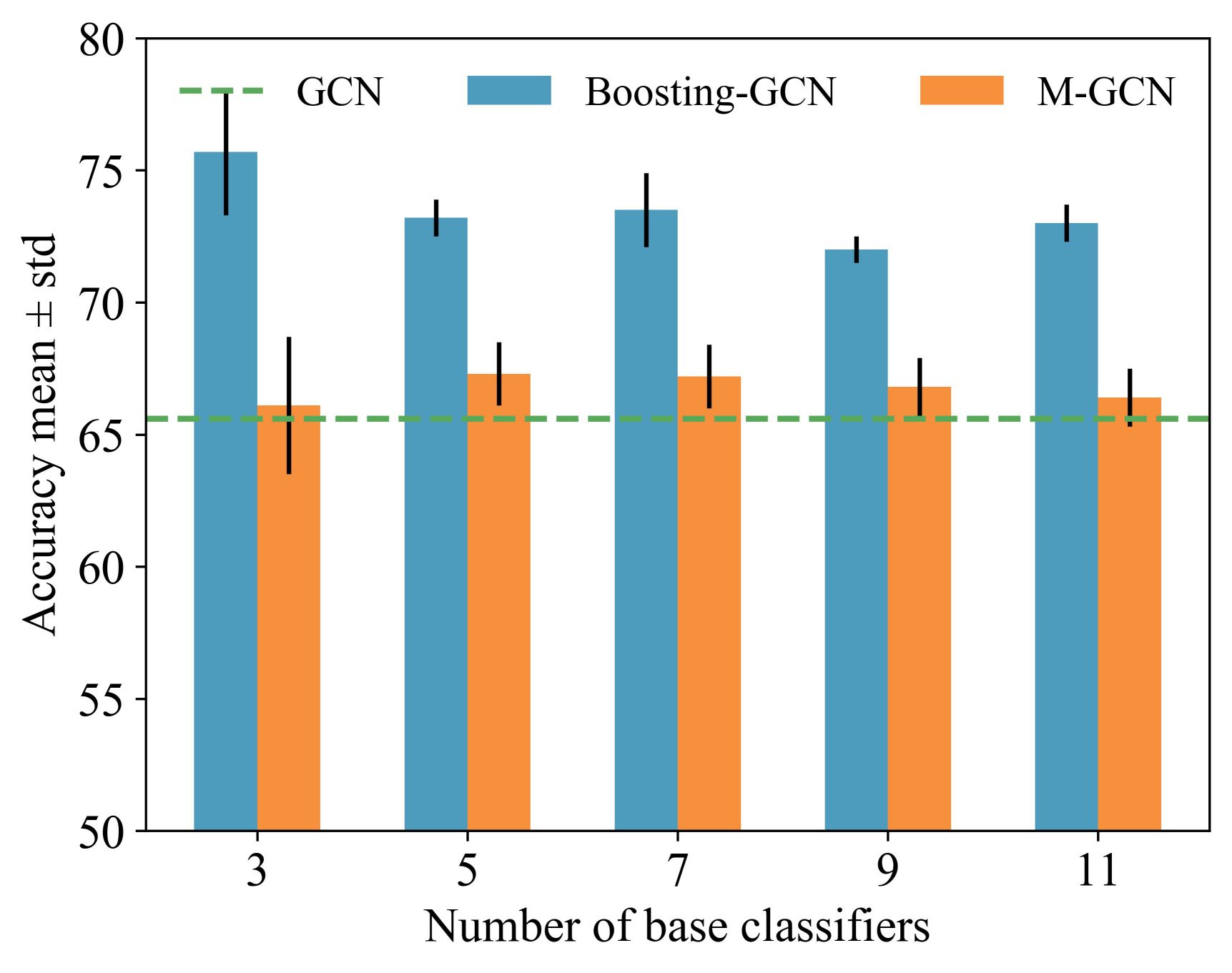}
\label{fig:4-1}
\end{minipage}%
}
\subfloat[Citeseer]{
\begin{minipage}[t]{0.3\linewidth}
\centering
\includegraphics[width=1.5in]{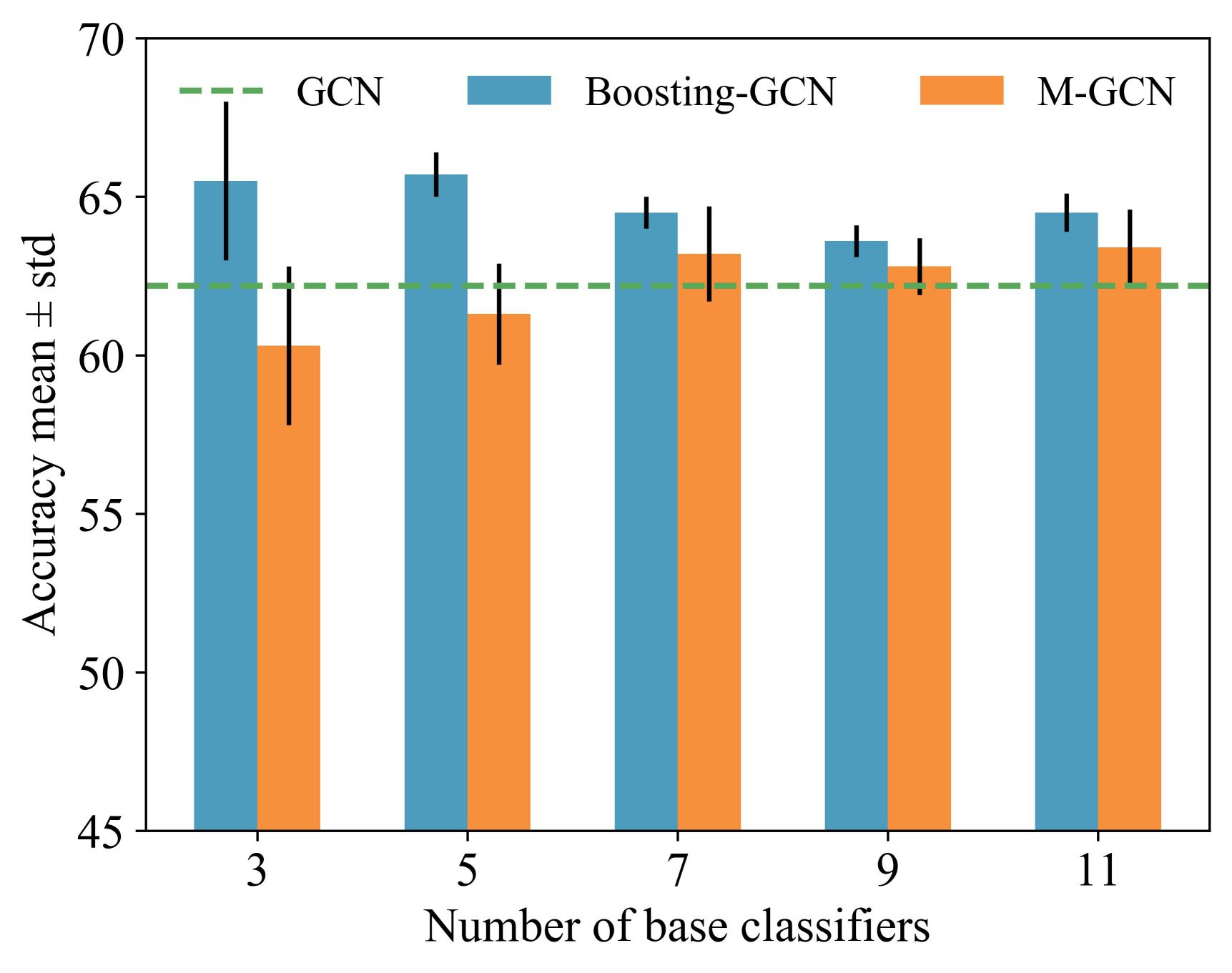}
\label{fig:4-2}
\end{minipage}
}
\subfloat[Pubmed]{
\begin{minipage}[t]{0.3\linewidth}
\centering
\includegraphics[width=1.5in]{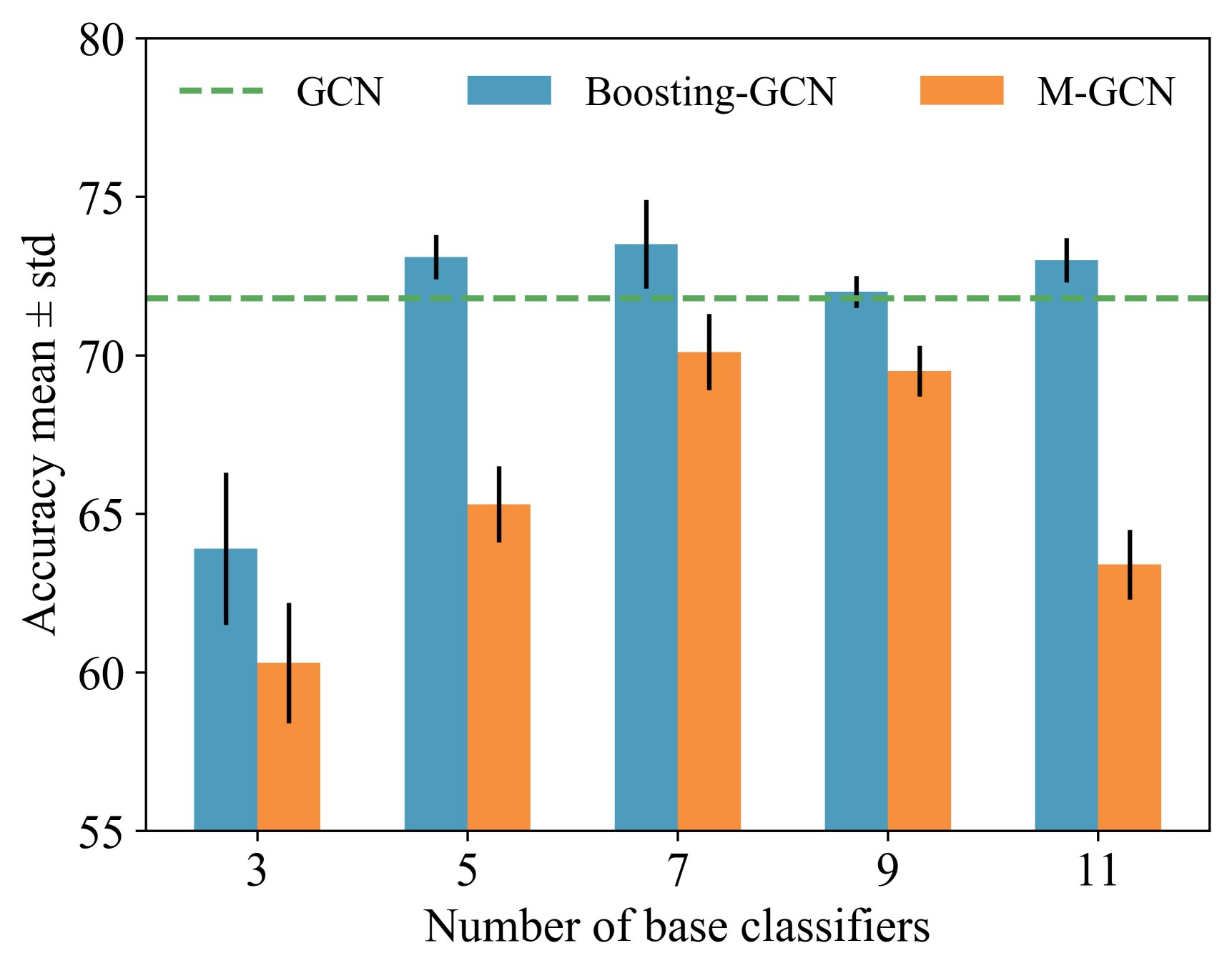}
\label{fig:4-3}
\end{minipage}%
}
\caption{Classification results of Boosting-GCN and M-GCN with different base classifiers.}
\label{fig:4}
\centering
\end{figure}

 We independently trained $M$ GCNs using the same strategy described in Equation (\ref{equ:test1}) and named this method M-GCN. We compare Boosting-GNN with M-GCN, which is trained according to the hard voting frameworks. Using the synthetic imbalanced datasets in Sect. \ref{sec:11}., we changed $M$ and conducted several experiments. Ten runs with different random seeds were conducted to calculate the mean and standard deviation. The experimental results are shown in Fig. \ref{fig:4}, and the classification results of GCN are represented by dotted lines. By effectively setting the number of base classifiers, Boosting-GCN significantly improves classification accuracy compared with M-GCN and GCN.

\par
\begin{figure}[h]
\centering
\subfloat[Boosting-GCN on Cora]{
\begin{minipage}[h]{0.3\linewidth}
\includegraphics[width=2in]{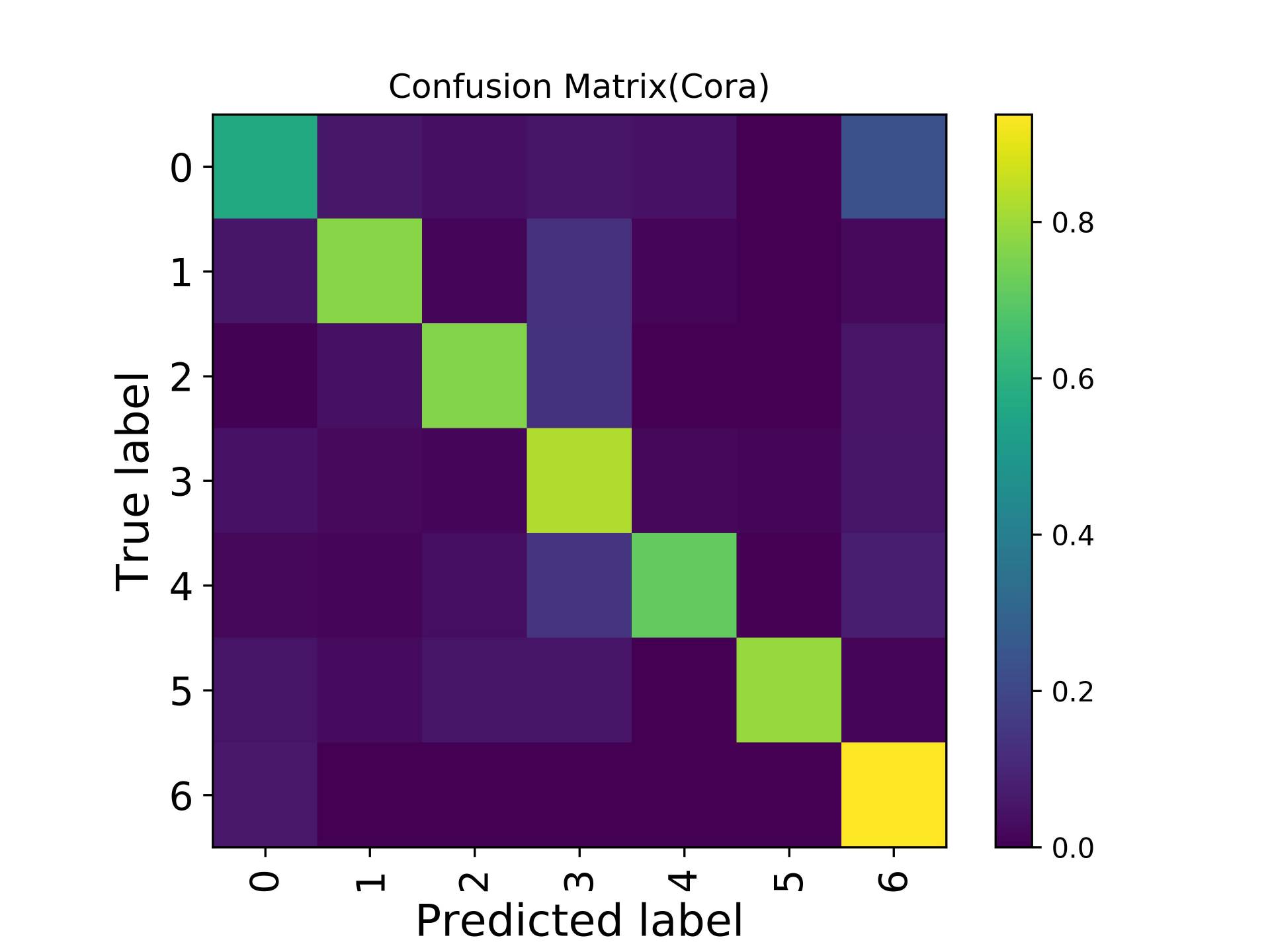}
\end{minipage}%
}
\subfloat[Boosting-GCN on Citeseer]{
\begin{minipage}[h]{0.3\linewidth}
\centering
\includegraphics[width=2in]{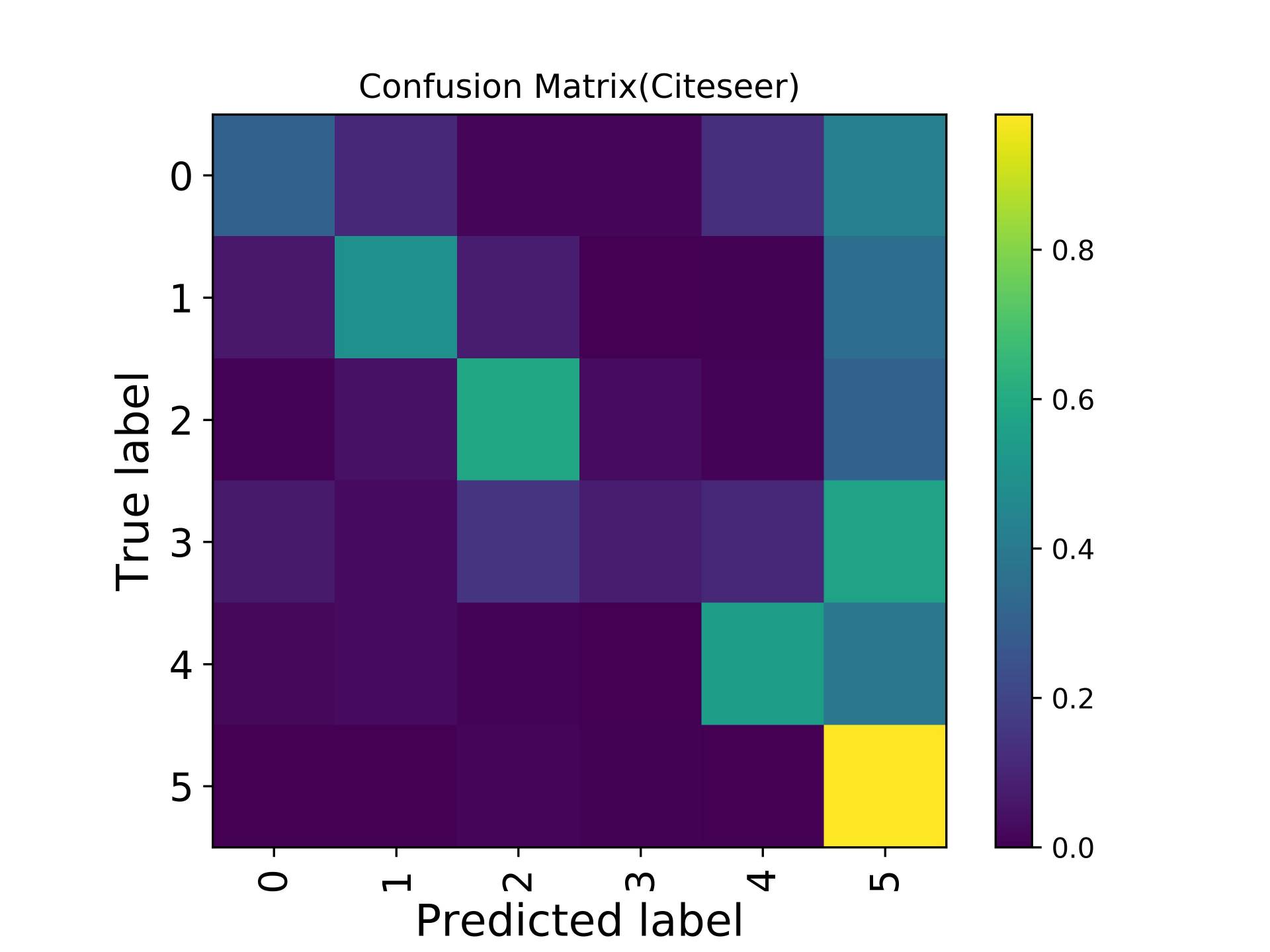}
\end{minipage}%
}
\subfloat [Boosting-GCN on Pubmed]{
\begin{minipage}[h]{0.3\linewidth}
\centering
\includegraphics[width=2in]{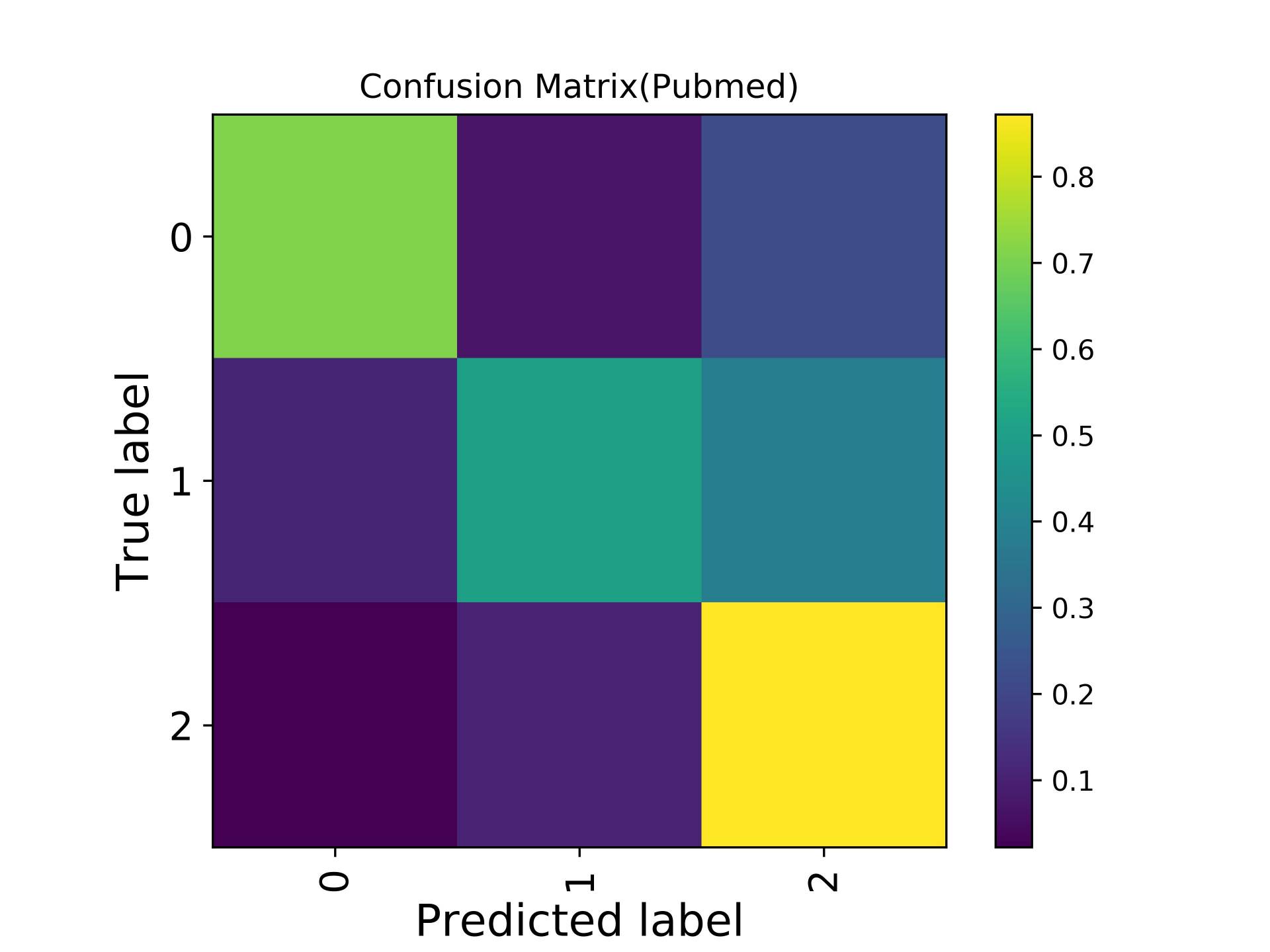}
\end{minipage}%
}

\subfloat[GCN on Cora]{
\begin{minipage}[h]{0.3\linewidth}
\centering
\includegraphics[width=2in]{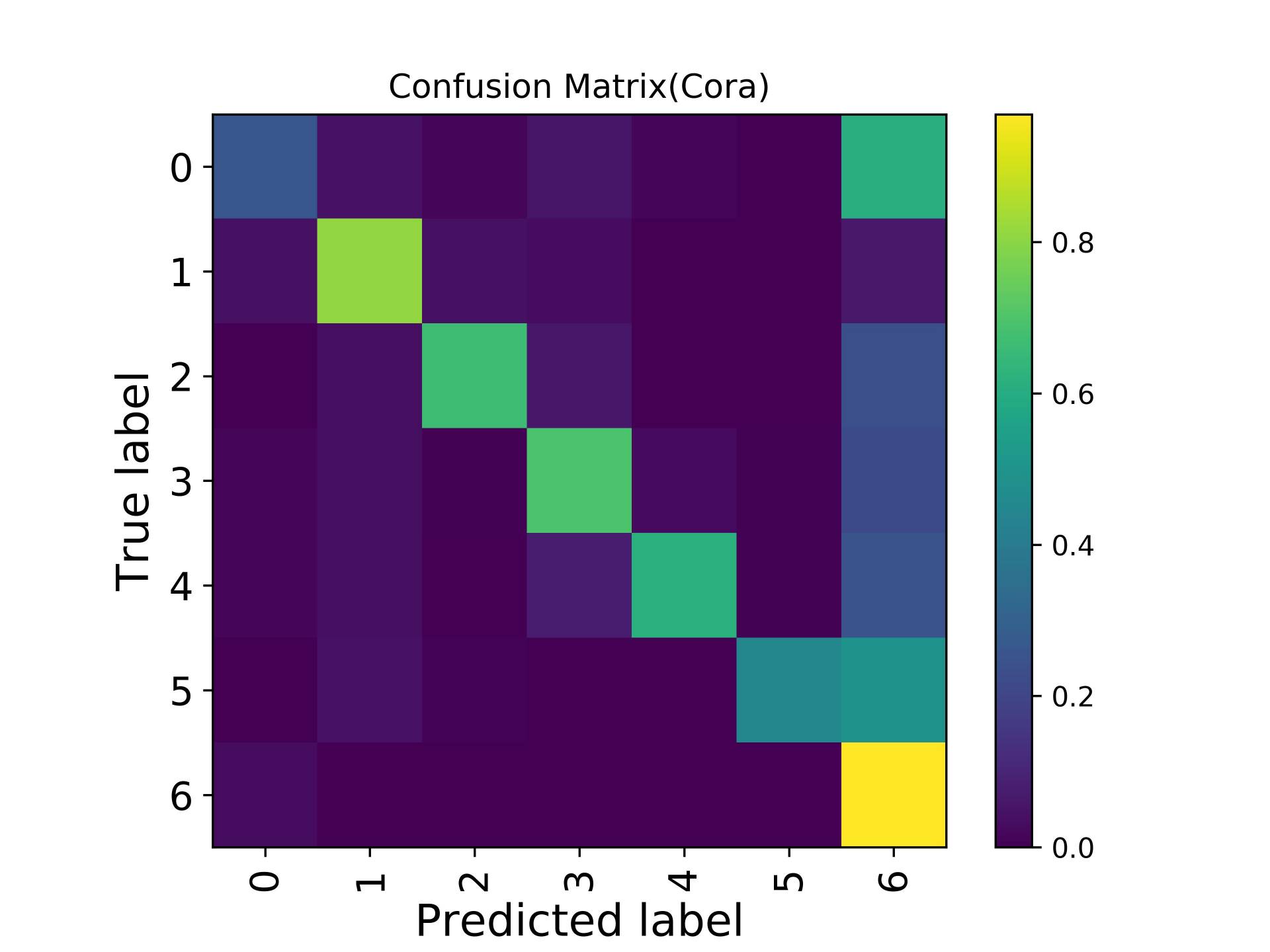}
\end{minipage}
}
\subfloat[GCN on Citeseer]{
\begin{minipage}[h]{0.3\linewidth}
\centering
\includegraphics[width=2in]{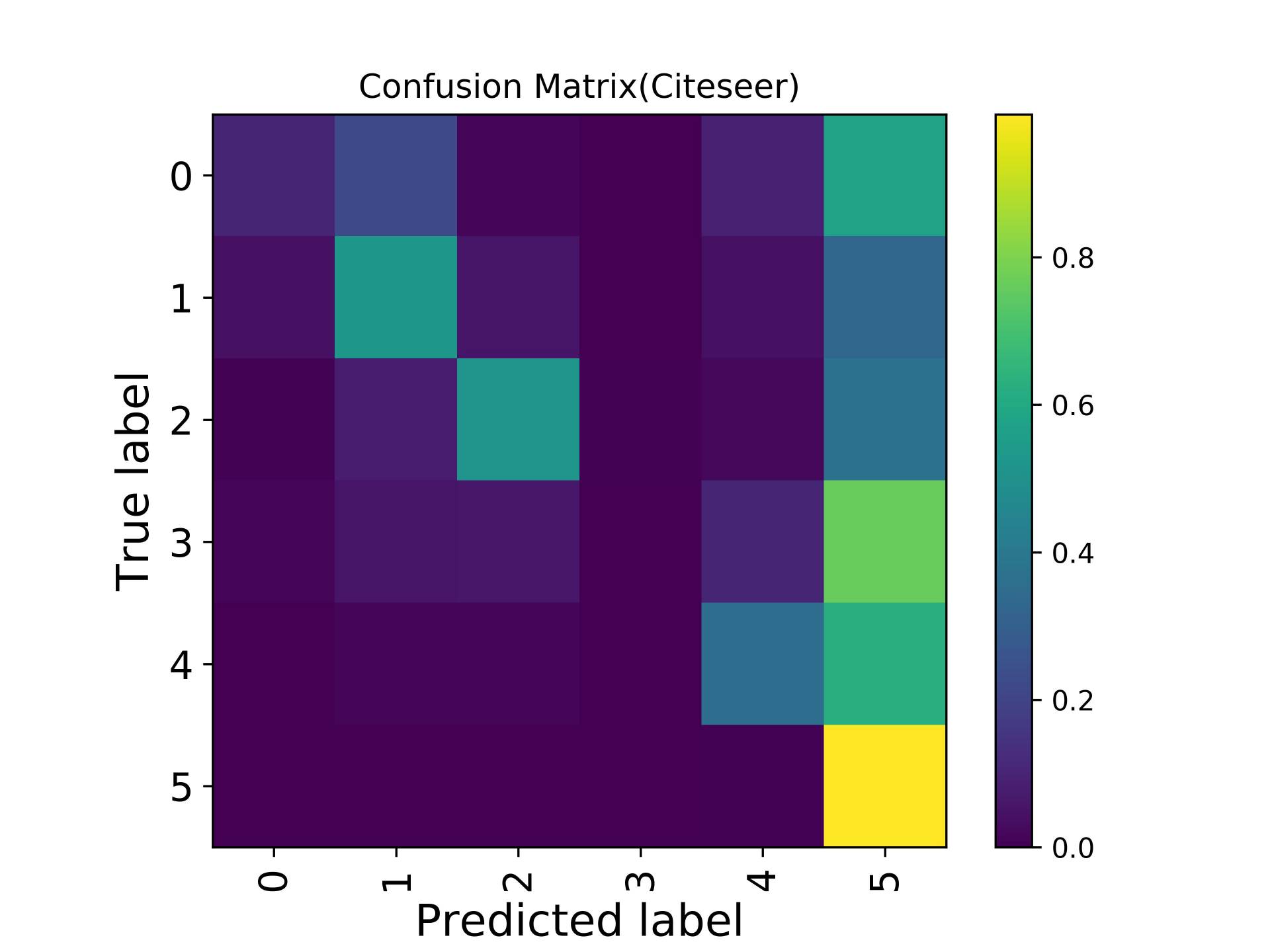}
\end{minipage}
}
\subfloat[GCN on Pubmed]{
\begin{minipage}[h]{0.3\linewidth}
\centering
\includegraphics[width=2in]{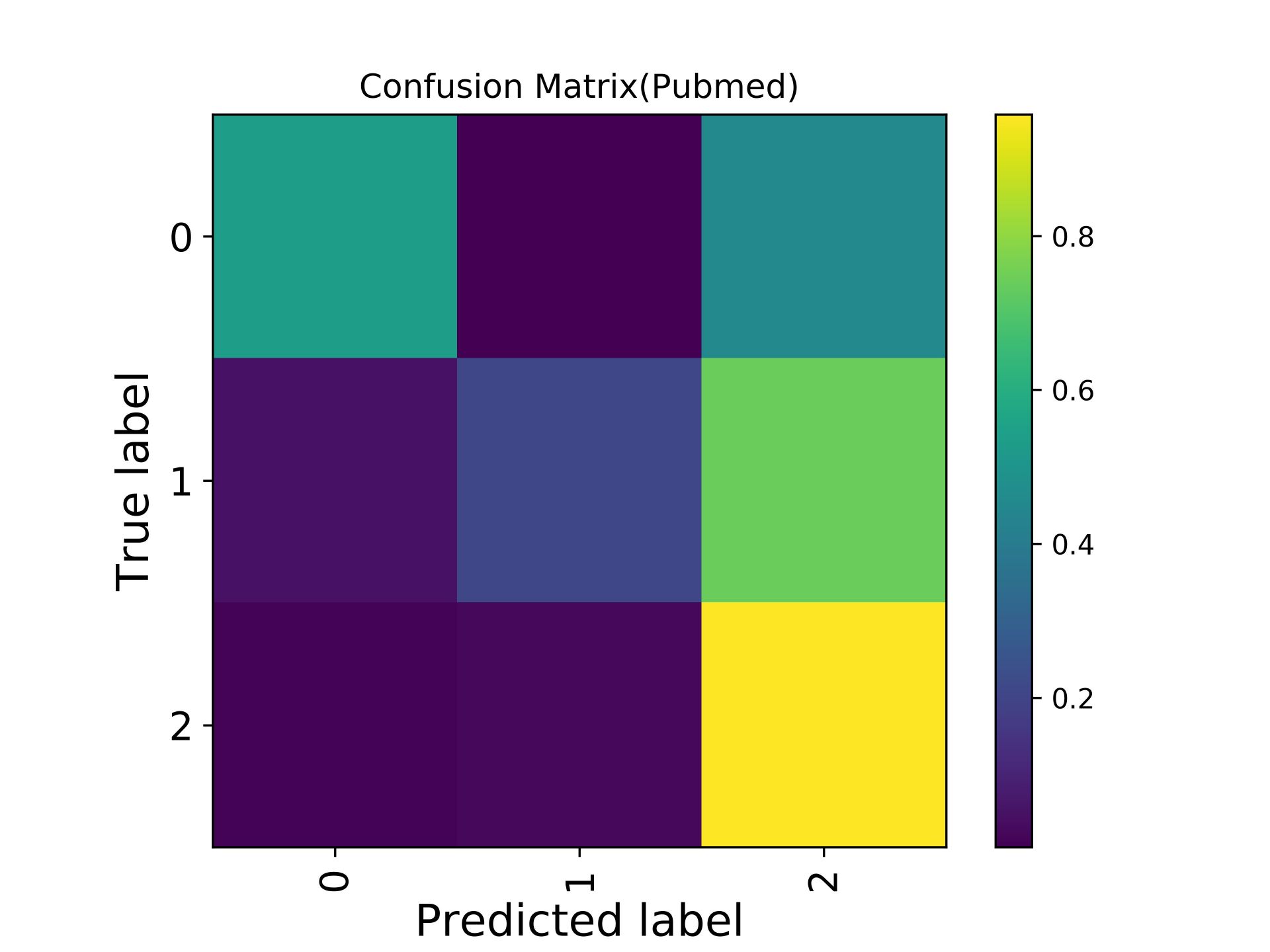}
\end{minipage}
}
\caption{Confusion matrix for the Cora, Pubmed and Citeseer datasets with 30 labeled nodes for majority class and 5 labeled nodes for minority class.}
\label{fig:5}
\centering
\end{figure}

 Next, in order to study the misclassification of samples, we observed the confusion matrix. To increase the imbalance, $s$ is set to 5. The last class is selected as the majority class, and the other classes are selected as the minority classes for convenience. Ten experiments are conducted, and the confusion matrix of the average experimental results is shown in Fig. \ref{fig:5}. Compared with the confusion matrix of the classification performed by GCN, Boosting-GCN achieves a better classification performance.

Due to the sample imbalance, the classifier tends to divide the samples into the majority class, which is reflected by the fact that the last column of the confusion matrix usually has the maximum value (with the brightest color). Compared with GNN, Boosting-GNN improves the performance to a certain extent, especially on Cora dataset. Based on the aggregation of base estimators, the values on the diagonal of the confusion matrix increase and the values in the last column of the confusion matrix decrease.
\par In summary, Boosting-GNN integrates multiple GNN classifiers to reduce the effect of overfitting to certain degree. Moreover, Boosting-GNN reduces the deviation caused by a single classifier and achieves better robustness. Boosting-GNN is an improvement of traditional GNN and makes AdaBoost compatible with GNN. Boosting-GNN achieves higher classification accuracy than a single GNN on imbalanced datasets with the same number of learning epochs.

\subsection{Analysis of Training Time}
\label{sec:16}
In this section, we conduct a time-consuming analysis of the experiment. We measure the training time on a NVIDIA Tesla V100 GPU. The time required to train the original GCN model for 100 epochs is 6.11s. The time consumed by M-GCN and Boosting-GCN is shown in the Table. Boosting-GCN-t and Boosting-GCN-w/o denote Boosting-GCN with transfer learning and Boosting-GCN without migration learning, respectively.

\begin{table}[h]
\centering
\caption{Comparison of running time when using different number of GCN base classifiers. We use Cora and train each base classifier for 100 epochs.}
\label{tab:4}       
\begin{tabular}{ccccc}
\hline\noalign{\smallskip}
Method &5-classifier &7-classifier &9-classifier\\
\noalign{\smallskip}\hline\noalign{\smallskip}
M-GCN             &28.76s &39.52s &51.04s \\
Boosting-GCN-t    &10.44s &13.43s &18.03s \\
Boosting-GCN-w/o  &18.36s &27.64s &34.83s \\
\noalign{\smallskip}\hline
\end{tabular}
\end{table}

Compared to GCN, Boosting-GCN consumes exponentially more time. However, Boosting-GCN reduces the training time by about 50\% compared to M-GCN. The application of transfer learning can significantly reduce the time consumed, and models can achieve similar accuracy.

\section{Conclusion}
\label{sec:17}
A multi-class AdaBoost for GNN, called Boosting-GNN, is proposed in this paper. In Boosting-GNN, a number of GNNs are used as base estimators, which are trained sequentially. Also, the errors of a previous GNN are used to update the weights of samples for the next GNN to improve performance. The weights of training samples are incorporated to the cross-entropy error function in the GNN back propagation learning algorithm. The appliance of transfer learning can significantly reduce the time consumed for computation. The performance of the proposed Boosting-GNN for processing imbalanced data is tested. The experimental results show that Boosting-GNN achieves better performance than state-of-the-arts on synthetic imbalanced datasets, with an average performance improvement of 4.5\%.



%
%

\bibliographystyle{apalike}
\bibliography{ref}
\end{document}